\documentclass[lettersize,journal]{IEEEtran}
\usepackage{caption}
\usepackage{amsmath,amsfonts,amsthm,amssymb}
\usepackage{algorithmic}
\usepackage{algorithm}
\usepackage{array}
\usepackage{textcomp}
\usepackage{stfloats}
\usepackage{url}
\usepackage{verbatim}
\usepackage{graphicx}
\usepackage{cite}
\usepackage{booktabs}
\usepackage{xcolor}
\usepackage{multirow}
\usepackage{colortbl}
\usepackage{adjustbox}
\usepackage{pifont}
\usepackage{stfloats}
\usepackage{subfigure}
\usepackage[backref]{hyperref}

\begin{document}

\makeatletter
\def\ps@IEEEtitlepagestyle{
  \def\@oddhead{\mycopyrightnotice}
  \def\@evenfoot{}
}
\def\mycopyrightnotice{
  {\footnotesize
  \begin{minipage}{\textwidth}
  \centering
  Copyright~\copyright~2026 IEEE. Personal use of this material is permitted. However, permission to use this  \\ 
  material for any other purposes must be obtained from the IEEE by sending a request to pubs-permissions@ieee.org.
  \end{minipage}
  }
}

\title{Adaptive Dual-Teacher Distillation with Subnetwork Rectification for Bridging Semantic Gaps in Black-Box Domain Adaptation}
\author{
Zhe Zhang, 
Jing Li,
Wanli Xue,
Xu Cheng,~\IEEEmembership{Senior Member,~IEEE},
Jianhua Zhang,~\IEEEmembership{Senior Member,~IEEE},
Qinghua Hu,~\IEEEmembership{Senior Member,~IEEE},
and Shengyong Chen,~\IEEEmembership{Senior Member,~IEEE}

\thanks{
This work was supported in part by 
Tianjin Natural Science Foundation Project (25JCQNJC00740, 23JCYBJC00360, 23JCJQJC00070), the Open Project of the Text Computing and Cognitive Intelligence Ministry of Education Engineering Research Center (No. TCCI250205), 
the Interdisciplinary Research Support Program for Young Faculty of Tianjin University of Technology (JCXK2026-01), and National Natural Science Foundation of China (62376197, 62673180, 62406224).
(\textit{Corresponding author: Jing Li.})

Zhe Zhang, Jing Li, Wanli Xue, Xu Cheng, Jianhua Zhang, and Shenyong Chen are with the School of Computer Science and Engineering, Tianjin University of Technology, Tianjin 300384, China, and also with the Engineering Research Center of Learning-Based Intelligent System, Ministry of Education of the People’s Republic of China, Tianjin University of Technology, Tianjin 300384, China. 
Jing Li is also with the Text Computing and Cognitive Intelligence Ministry of Education Engineering Research Center, Guizhou University, Guiyang 550025, China 
(e-mail: zzhe0124@stud.tjut.edu.cn; 
jing\underline{ }li@tju.edu.cn;
xuewanli@email.tjut.edu.cn;
xu.cheng@ieee.org;
zjh@ieee.org;
sy@ieee.org).

Qinghua Hu is with the School of Artificial Intelligence, Tianjin University, Tianjin 300350, China, also with the Engineering Research Center of City Intelligence and Digital Governance, Ministry of Education of the People’s Republic of China, Tianjin University, Tianjin 300350, China, and also with the Haihe Laboratory of Information Technology Application Innovation, Tianjin 300051, China (e-mail: huqinghua@tju.edu.cn).}
}

\maketitle

\begin{abstract}
Assuming that neither source data nor source-model parameters are accessible, black-box domain adaptation (BBDA) represents a highly practical yet challenging setting, where transferable knowledge is limited to the predictions of a black-box source model. Existing approaches exploit such knowledge via pseudo-label refinement or by leveraging vision-language models (ViLs), but they often fail to reconcile the inherent discrepancy between task-specific knowledge from black-box models and language-aligned semantic priors of ViLs, resulting in suboptimal integration and degraded adaptation performance. To address this challenge, we propose adaptive Dual-Teacher Distillation with Subnetwork Rectification (DDSR), a framework that explicitly reconciles these complementary yet inconsistent knowledge sources. DDSR employs an adaptive prediction fusion strategy to integrate predictions from the black-box source model and a ViL, generating reliable pseudo-labels for the target domain. A subnetwork-based regularization mechanism mitigates overfitting to noisy supervision by enforcing output consistency and gradient divergence. Furthermore, progressively improved target predictions iteratively refine both pseudo-labels and ViL prompts, enhancing semantic alignment. Finally, class-wise prototypes are used to further optimize target predictions via self-training. {Extensive experiments on multiple benchmarks demonstrate that DDSR achieves state-of-the-art performance among existing BBDA methods and remains competitive with many approaches that have access to source data or source-model parameters.}
\end{abstract}


\section{Introduction}
\IEEEPARstart{U}{nsupervised} domain adaptation (UDA) \cite{zhang2022transfer,RoadDamageDetection2026csvt} addresses the challenge of distribution shifts when transferring knowledge from a labeled source domain to an unlabeled target domain. Conventional UDA methods assume simultaneous access to both source and target data. However, this assumption is often unrealistic in practice, as source data are frequently subject to privacy restrictions or cannot be shared. To mitigate this limitation, source-free domain adaptation (SFDA) \cite{liang2020we,ASOGEcsvt2024SFDA,liupanDPL} relies solely on a source model pre-trained with labeled source data and adapts it to the unlabeled target domain. Nevertheless, since raw source samples can potentially be reconstructed through generative techniques \cite{goodfellow2014generative} when the source model is available, the risk of private information leakage remains.

To address this issue, black-box domain adaptation (BBDA) \cite{liang2022dine} treats the source model as a black-box predictor, transferring knowledge to the target domain solely through the predictions it produces on target samples. In this setting, both the source data and the internal details of the source model, such as its architecture and parameters, are inaccessible, which makes BBDA highly challenging. Beyond preserving privacy, BBDA does not require the target model to share the same structure as the source model, since the latter’s design is unknown. This flexibility enables effective adaptation even on resource-constrained devices. Moreover, as artificial intelligence services delivered via APIs become increasingly popular \cite{achiam2023gpt}, the BBDA paradigm, where the target model queries the source model for predictions, aligns well with this trend. Considering these factors, BBDA is both important and holds significant potential for future development.

Compared to UDA and SFDA, research on BBDA remains relatively limited. Due to the stricter constraints in the BBDA setting, existing UDA and SFDA methods cannot be directly applied. Distribution shifts between the source and target domains often cause the black-box source model to produce inaccurate predictions on target samples. To maximize the utility of target data, most BBDA methods focus on reducing prediction noise through techniques such as knowledge distillation \cite{liang2022dine,hinton2015distilling}, consistency learning \cite{DBLP:conf/bmvc/ZhangZJ021}, and feature regularization \cite{DBLP:conf/ijcai/PengDL0023}. However, without high-level semantic supervision, the performance of these data-driven approaches remains limited. Beyond exploiting black-box predictions, other works \cite{xiao2024adversarial,tian2024clip} incorporate the vision-language model (ViL) named CLIP \cite{radford2021learning} in target model training. Benefiting from diverse training data and semantically driven objectives, ViLs capture general semantic knowledge beyond pixel-level or local features, making them more robust to distribution shifts. 

Nevertheless, directly leveraging ViLs in the BBDA setting is non-trivial. In particular, the knowledge encoded in ViLs is often inconsistent with that of the black-box source model in terms of decision boundaries and semantic granularity \cite{xiao2024adversarial}. For example, ViLs tend to produce semantically smoother predictions, whereas the black-box model may exhibit domain-specific biases and sharper class boundaries. Such discrepancies can give rise to conflicting supervisory signals when both sources are used to guide the target model. Notably, prior work \cite{xiao2024adversarial} has empirically shown that directly combining the predictions from ViLs and black-box models (e.g., via naive averaging) does not yield optimal performance in the same BBDA setting, suggesting that this inconsistency can negatively affect the adaptation process. Therefore, effectively reconciling the discrepancy between ViLs and black-box models requires a principled mechanism to balance their complementary strengths while mitigating conflicting supervision.

Inspired by prior works \cite{xiao2024adversarial,tian2024clip}, we propose a dual-teacher distillation with subnetwork rectification (DDSR) for BBDA. To address the challenge of reconciling task-specific knowledge from black-box source models with language-aligned semantic priors from vision–language models, DDSR is structured in two stages. In the first stage, the black-box source model and CLIP act as dual teachers. Target samples are fed into both teachers to obtain predictions, which are integrated by an adaptive prediction fusion module. This module is designed to reconcile their complementary yet potentially inconsistent predictions, producing reliable pseudo-labels to supervise the target network.

To mitigate overfitting to noisy pseudo-labels arising from imperfect fusion and domain shift, we introduce a subnetwork that shares part of the structure and parameters with the target network. This subnetwork rectifies the target network through output alignment and gradient divergence. As training progresses, the target model produces increasingly reliable predictions, which are leveraged to iteratively refine pseudo-labels and update CLIP prompts, enabling both to better adapt to the target domain.

In the second stage, target features and their corresponding class predictions are extracted via the target model. Subsequently, class-wise prototypes are constructed based on the feature representations within each category. Finally, target labels are rectified by aligning them with their nearest prototypes, and the refined pseudo-labels are used to further optimize the target model. 

Our contributions can be summarized as follows:
\begin{itemize}
\item 
To jointly exploit the task-specific knowledge of black-box source models and the language-aligned semantic priors of CLIP, we propose an adaptive prediction fusion strategy that reconciles their complementary yet inconsistent predictions to generate reliable pseudo-labels. To mitigate overfitting to noisy supervision, we further introduce a subnetwork to regularize the target model via output consistency and gradient divergence.  
\item 
We further develop a progressive self-distillation mechanism that leverages the improving predictive capability of the target model to iteratively refine pseudo-labels and CLIP prompts, enabling better semantic alignment with the target domain. In addition, class-wise prototypes are introduced to rectify target predictions, facilitating more accurate and semantically coherent pseudo-labels.
\item {Extensive experiments on multiple benchmarks validate the effectiveness of DDSR, which achieves state-of-the-art performance among existing BBDA methods and remains competitive with several UDA and SFDA methods that have access to source data or source-model parameters.} 
\end{itemize}

The remainder of this paper is organized as follows. Section II reviews the related work. Section III presents the problem setting and details the proposed method. Section IV reports the experimental results and analysis on multiple benchmarks. Finally, Section V concludes the paper.

\section{Related Work}
\subsection{Domain Adaptation} 
To mitigate the distributional discrepancy between the two domains, unsupervised domain adaptation (UDA) approaches have evolved along several major directions, including feature alignment \cite{long2015learning,long2016unsupervised}, adversarial learning \cite{ganin2016domain,TMM23RADAprompt,csvt23WDAN}, and self-supervised representation learning \cite{TMM2024Tadapter,french2018selfensembling,kang2019contrastive,he2023independent,csvt24Neno}. Despite their effectiveness, these methods require direct access to source data, which limits their applicability in privacy-sensitive or proprietary scenarios. Source-free domain adaptation (SFDA) \cite{liupanDPL,peng2026lifelong,PCTSCcsvt2026SFDA} addresses this limitation by assuming that the source data are inaccessible during adaptation, while the pre-trained architecture and parameters of the source model {are} available. Most SFDA methods rely on pseudo-label self-training as the core strategy. A representative example, SHOT \cite{liang2020we}, decouples the feature extractor and classifier from the source model and optimizes the target model via entropy minimization and information maximization. Subsequent studies improve upon this by incorporating consistency regularization \cite{Tang_2023_ICCV} or class-wise prototype alignment \cite{ZHOUClassPrototypeDiscovery,Qiu2021CPGA}. Although SFDA effectively removes the need for source data, it still {relies} on access to the source-model parameters, which may pose potential privacy risks and hinder broader deployment \cite{liang2022dine}.

To enhance semantic supervision in SFDA, recent works have incorporated vision-language models (ViLs) . For example, Co-learn++ \cite{zhang2025source} employs a dual-branch collaborative learning strategy, where the target model and a pre-trained ViL model iteratively refine each other's predictions to correct the bias of the source model in the target domain. Similarly, ProDE \cite{Tang2025ProDE} uses prompt-based adaptation of pre-trained ViLs to inject language-aligned semantic priors, generating more reliable pseudo-labels for self-training in the target domain. While these ViL-assisted SFDA methods improve semantic guidance and adaptation performance, they still assume access to source-model parameters and are not designed for scenarios where parameters of the source models are unavailable. This motivates the need for approaches, such as our DDSR framework, that can effectively reconcile predictions from a black-box source model with external semantic priors, generating reliable pseudo-labels and robustly training the target model.

\subsection{Black-box Domain Adaptation}
Black-box Domain Adaptation (BBDA) further relaxes the source accessibility assumption by prohibiting access to both source data and source-model parameters, while allowing only query access to the predictions of the source model. This setting better reflects real-world scenarios, such as cross-institutional collaboration and commercial model deployment, where data sharing and model exposure are restricted. However, BBDA poses unique challenges in addressing domain shift and mitigating pseudo-label noise, as transferable information is limited to model outputs. Early BBDA methods, such as DINE \cite{liang2022dine}, employ entropy-based self-training to iteratively refine pseudo-labels using black-box predictions. Subsequent works extend this paradigm by exploring distribution calibration \cite{zhang2024reviewing}, feature separation \cite{xia2024separation}, memorization mechanisms \cite{zhang2023black}, and regularization-based strategies \cite{DBLP:conf/ijcai/PengDL0023}. Despite their effectiveness, these approaches predominantly rely on predictions from black-box models trained on specific tasks, which provide limited semantic information beyond the predefined label space \cite{xiao2024adversarial}. Such supervision may be insufficient to capture richer semantic relationships, particularly under significant domain shifts. As observed in prior work \cite{xiao2024adversarial}, this limitation may lead to suboptimal adaptation performance, where the target model exhibits reduced discriminability and generalization ability in challenging scenarios.

To address this limitation, recent studies have explored incorporating complementary semantic knowledge from ViLs \cite{Zhang_2024PAMI_VLMs}. Models such as CLIP \cite{radford2021learning} provide language-aligned representations learned from large-scale pretraining, offering richer semantic priors beyond closed-set label spaces. Several BBDA approaches \cite{xiao2024adversarial,tian2024clip} leverage CLIP to enhance pseudo-label quality and mitigate noisy supervision. For example, AEM \cite{xiao2024adversarial} introduces dual-branch supervision with adversarial alignment to integrate knowledge from both models, while BBC \cite{tian2024clip} dynamically selects between predictions from the source model and CLIP based on confidence scores.
Nevertheless, effectively combining task-specific knowledge from black-box models with the general semantic priors of ViLs remains non-trivial, due to their inherent discrepancies in prediction behavior and semantic granularity. This motivates the need for a principled mechanism, such as our DDSR framework, to reconcile these complementary sources and achieve reliable pseudo-label generation.

{BBC and AEM are closely related to our method in that they also introduce vision-language models to complement the limited knowledge available from a black-box source model. Nevertheless, the three methods differ substantially in how heterogeneous knowledge is utilized and adapted. BBC estimates prediction confidence based on entropy and selects either the black-box or CLIP prediction for each target sample, followed by neighborhood-based pseudo-label refinement. In contrast, our DDSR performs reliability-aware soft fusion, preserving the complementary information from both teachers and adaptively adjusting their contributions at the supervision level. AEM adopts a shared feature extractor with two classifiers separately guided by the black-box and ViL experts, and employs adversarial discrepancy optimization to align the heterogeneous expert knowledge. DDSR instead first integrates the two teachers into unified supervision and subsequently employs a parameter-sharing subnetwork to rectify target learning through output consistency and gradient divergence, thereby mitigating overfitting to noisy pseudo-labels. Moreover, while BBC and AEM also progressively update supervision or expert knowledge during adaptation, DDSR couples dual-teacher fusion and subnetwork rectification with target-knowledge feedback, where the emerging target model progressively refines the fused pseudo-labels and ViL prompts before prototype-based self-training further consolidates the target-domain class structure.}
 
\subsection{Vision-language Model}
Vision–language models (ViLs) aim to establish semantic associations between visual and linguistic modalities, enabling cross-modal understanding. A representative ViL, CLIP \cite{radford2021learning}, achieves effective alignment between images and text via large-scale contrastive learning on over 400 million image–text pairs, offering strong zero-shot generalization and rich semantic knowledge. Other foundational ViLs have been proposed, such as BLIP \cite{li2022blip}, which focuses on pretraining with image–text pairs and captioning objectives, OpenCLIP \cite{cherti2023reproducible}, which provides open-source implementations of CLIP-style models with diverse pretraining datasets, and ALIGN \cite{jia2021scaling}, trained on billions of noisy image–text pairs to scale cross-modal representations. In our work, we adopt CLIP to leverage its strong zero-shot capability and mature implementation for pseudo-label generation. Nevertheless, our framework is flexible and could alternatively incorporate other ViLs to provide complementary semantic supervision in future extensions.

\begin{figure*}[!t]
\centering
\includegraphics[width=1\textwidth]{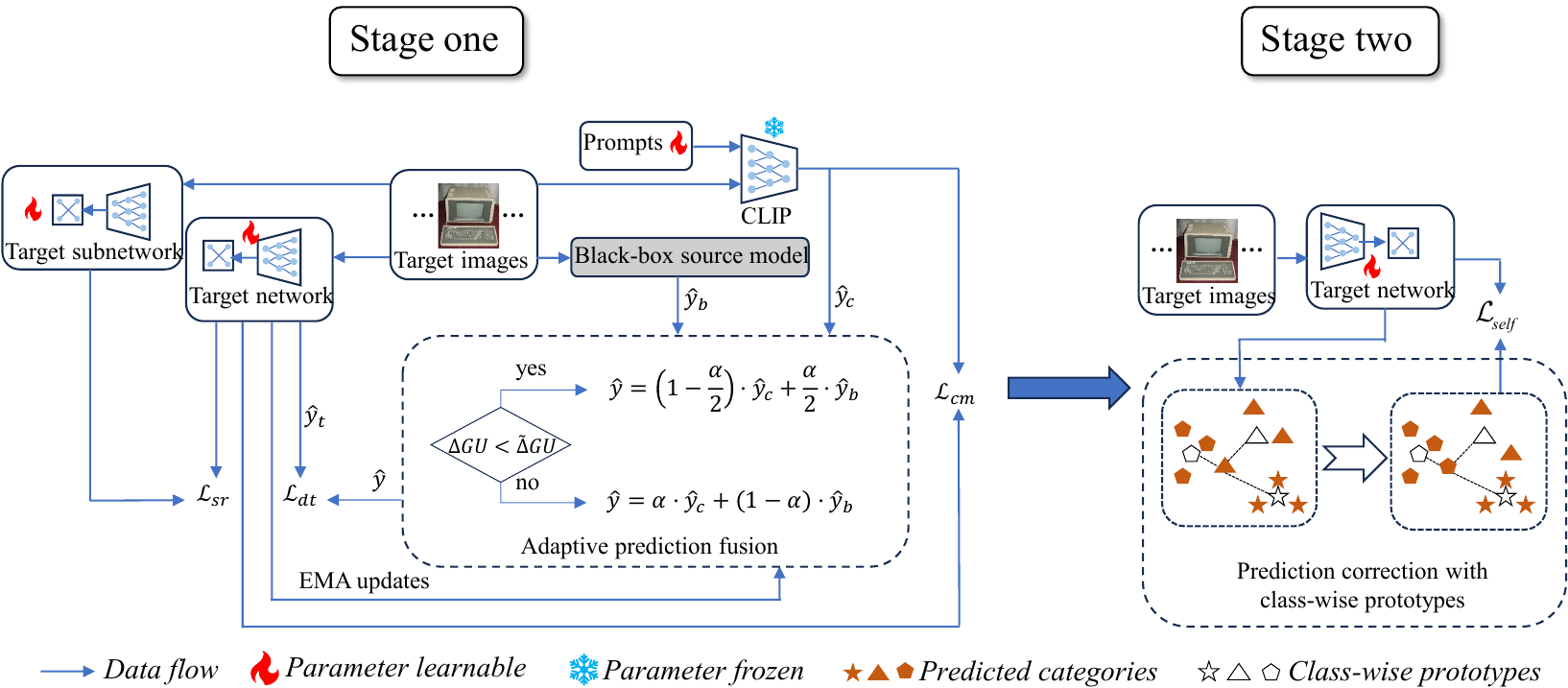}
\caption{The overview of our proposed DDSR framework. The training process consists of two stages. In Stage one, DDSR adaptively integrates the complementary predictions from the black-box source model and CLIP to generate reliable pseudo-labels for the target domain. A subnetwork-driven regularization strategy is introduced to alleviate overfitting caused by noisy supervision. The target predictions are further employed to iteratively update pseudo-labels through exponential moving average (EMA) and to learn the ViL prompts via the loss $\mathcal{L}_{cm}$. In Stage two, the target model is further optimized through self-training based on class-wise prototypes, leading to more discriminative and semantically consistent feature representations.}
\label{overview}
\end{figure*}

\section{Methodology}
\subsection{Preliminaries}
In the BBDA setting, neither the labeled source data nor the parameters of the pre-trained source model $f_{s}$ are accessible during the target model training. Source knowledge is available only through the class predictions produced by the black-box source model on target samples. Consequently, the target model is initialized independently, without inheriting any weights from the source model, ensuring that the adaptation strictly follows the BBDA protocol. The unavailable labeled data of the source domain is denoted as $D_{s}=\{x_{s}^{i},y_{s}^{i}\}_{i=1}^{n_{s}}$, where $n_{s}$ is the total number of samples in the source domain, and $x_{s}^{i} \in \mathcal{X}_{s}$, $y_{s}^{i} \in \mathcal{Y}_{s}$. The {unlabeled} data of the target domain is denoted as $D_{t}=\{x_{t}^{i}\}_{i=1}^{n_{t}}$, where $n_{t}$ is the total number of samples in the target domain, and $x_{t}^{i} \in \mathcal{X}_{t}$. Given such constrained source information, the goal of BBDA is to train a robust target model $f_{t} = g_{t} \circ h_{t}$ that can accurately predict $y_{t}^{i} \in \mathcal{Y}_{t}$ for target samples, where $g_{t}$ and $h_{t}$ denote the feature extractor and classifier, respectively. 
The source and target domains share the same label space $\mathcal{Y}_{s} = \mathcal{Y}_{t}$, where $|\mathcal{Y}_{t}| = C$. 

As shown in Fig.\ref{overview}, our framework consists of two stages. In Stage One, we exploit both the knowledge from the black-box source model and the semantic information provided by CLIP. Specifically, the entropy of the predictions from the two models is leveraged to adaptively determine their weights in the predictive fusion module, producing fused predictions that serve as noisy pseudo-labels for distilling knowledge into the target model. To reduce overfitting to noisy pseudo-labels, a subnetwork is initialized with partial parameters of the target model and employed to regularize the training of the target model. Moreover, predictions from the target model are leveraged to progressively refine the pseudo-labels and learn the CLIP prompts, injecting target-specific information into the supervision. In Stage Two, further refinement is performed by computing class-wise prototypes and reassigning each sample to the class of its nearest prototype. The corrected predictions are then used to train the target model, yielding additional performance improvements.
\subsection{Stage One}
\subsubsection{Dual-Teacher Knowledge Distillation} \label{FusionStrategy}
Knowledge distillation (KD) \cite{hinton2015distilling} facilitates the transfer of knowledge from a high-performing teacher model to a student model. Since KD does not depend on the internal structure or parameters of the source model, it is particularly suitable for the BBDA setting, where the source model serves as the teacher and the target model as the student. Although the source and target domains share the same label space, substantial distribution shifts often exist between them, which can cause the source model to produce noisy predictions on target samples. To mitigate this issue, we incorporate CLIP \cite{radford2021learning} as an additional teacher. Trained on large-scale image-text paired data, CLIP provides strong semantic understanding and cross-domain generalization, thereby enhancing the reliability of the pseudo-labels and improving the robustness of KD. 
  
We obtain the label predictions of the target samples $x_t$ from both the black-box source model and CLIP. Let $\hat{y}_{b}$ denote the soft predictions generated by the source model, and $\hat{y}_{c}$ denote those from CLIP. 
The fusion of these two predictions serves as the teacher signal for knowledge distillation. We find the fusion ratio between $\hat{y}_{b}$ and $\hat{y}_{c}$ plays a critical role in determining the effectiveness of distillation. Rather than manually specifying this ratio, we propose an adaptive prediction fusion strategy based on uncertainty estimates at both the instance level and the dataset level for the black-box source model and CLIP on the target domain.

We first quantify the instance-level predictive uncertainty using the Shannon entropy of the predictive distribution for each target sample $x_t^i$. Specifically, for a given sample, the uncertainty of the black-box source model and CLIP is computed as:
\begin{equation}
\label{H_b}
H(\hat{y}^i_b) = -\sum_{j=1}^{C} (\hat{y}^i_b)_j \log (\hat{y}^i_b)_j,
\end{equation}
\begin{equation}
\label{H_c}
H(\hat{y}^i_c) = -\sum_{j=1}^{C} (\hat{y}^i_c)_j \log (\hat{y}^i_c)_j.
\end{equation}
We then define the average instance-level uncertainty, referred to as individual uncertainty (IU), as:
\begin{equation}
IU_b = \mathbb{E}_{x^i_{t} \in \mathcal{X}_{t}} \left[ H(\hat{y}^i_b) \right], \quad
IU_c = \mathbb{E}_{x^i_{t} \in \mathcal{X}_{t}} \left[ H(\hat{y}^i_c) \right].
\end{equation}
A lower IU indicates more confident predictions on individual target samples.

To characterize the prediction behavior at the dataset level, we further define the global uncertainty (GU) as the entropy of the marginal predictive distribution:
\begin{equation}
GU_b = -\sum_{j=1}^{C} \left( \frac{1}{n_t} \sum_{i=1}^{n_t} (\hat{y}^i_{b})_j \right)
\log \left( \frac{1}{n_t} \sum_{i=1}^{n_t} (\hat{y}^i_{b})_j \right),
\end{equation}
\begin{equation}
GU_c = -\sum_{j=1}^{C} \left( \frac{1}{n_t} \sum_{i=1}^{n_t} (\hat{y}^i_{c})_j \right)
\log \left( \frac{1}{n_t} \sum_{i=1}^{n_t} (\hat{y}^i_{c})_j \right).
\end{equation}
GU reflects the diversity of predictions across the target set, where a higher value indicates reduced risk of collapsing into a limited subset of categories.

Based on IU and GU, the adaptive prediction fusion is formulated as:
\begin{equation}
\hat{y} =
\begin{cases}
\left(1 - \frac{\alpha}{2}\right) \hat{y}_{c} + \frac{\alpha}{2} \hat{y}_{b}, & \text{if} \Delta GU < \tilde{\Delta} GU, \\
\alpha \hat{y}_{c} + (1 - \alpha) \hat{y}_{b}, & \text{otherwise},
\end{cases}
\label{quanzhong}
\end{equation}
where $\alpha = IU_c/(IU_b + IU_c)$, $\Delta GU = GU_{b} - GU_{c}$, and $\tilde{\Delta} GU$ is a predefined threshold.

{From an information-theoretic perspective, IU and GU characterize complementary properties of the teacher predictions: IU reflects instance-level uncertainty, whereas GU reflects the diversity of marginal predictions over the target domain. Accordingly, Eq.~(6) employs two weighting regimes based on the global uncertainty discrepancy $\Delta GU$. When $\Delta GU < \tilde{\Delta}GU$, the diversity discrepancy is limited, and the fusion primarily exploits instance-level confidence, assigning a larger weight to CLIP as its relative confidence increases. In contrast, when $\Delta GU \geq \tilde{\Delta}GU$, the substantial diversity discrepancy suggests that confident CLIP predictions may concentrate on a limited subset of classes. The second branch therefore reduces the CLIP contribution as its relative confidence increases, retaining more source-model predictions to counterbalance such concentration. Thus, the different roles of $\alpha$ in the two branches are intentional, reflecting the trade-off between instance-level confidence and dataset-level prediction diversity.}

The fused predictions $\hat{y}$ encapsulate the knowledge from both teachers. To transfer this knowledge, we minimize the Kullback–Leibler (KL) divergence between the target model’s predictions $\hat{y}_{t}$ and $\hat{y}$, thereby guiding the target model to learn from the dual teachers while adapting to the target domain. The knowledge distillation loss is defined as:
\begin{equation}
\mathcal{L}_{kd} = \mathbb{E}_{x^i_t \in \mathcal{X}_t}  D_{KL} \left( \hat{y}^i \parallel \hat{y}^i_t \right),
\label{losskd}
\end{equation}
where $D_{KL}$ denotes the KL divergence, $\hat{y}^i_t = f_t(x^i_t)$, and $\hat{y}^i$ works as a pseudo-label for a target sample $x^i_t$. 
 
Given the source information is highly constrained, the underlying structure of the target data should also be exploited. 
Specifically, we employ a Mixup-based consistency loss \cite{TMM23DSmix} to improve the robustness and generalization of the target model. By linearly interpolating two randomly selected samples along with their corresponding predictions, the model is regularized to produce consistent outputs for the mixed inputs. 
The details are as follows:

\begin{align}
\mathcal{L}_{mix}
&= \mathbb{E}_{x_t^i, x_t^j \in \mathcal{X}_t}\;
   \mathbb{E}_{\lambda \sim \mathrm{Beta}(0.3,0.3)} \nonumber \\[4pt]
&\quad D_{KL}\!\left(
f_t(\mathrm{Mix}_\lambda(x_t^i, x_t^j))
\;\middle\|\;
\mathrm{Mix}_\lambda(\hat{y}_t^i, \hat{y}_t^j)
\right),
\label{lossmix}
\end{align}

where $f_t$ denotes the target model, $\mathrm{Mix}_\lambda (a,b) = \lambda\cdot a+(1-\lambda)\cdot b $ represents the Mixup operation.

The collapse of the target model is alleviated by the information maximization loss \cite{liang2020we} whose definition is as follows,
\begin{equation}
\begin{aligned}
\mathcal{L}_{im} = H\left(\mathbb{E}_{x^i_{t} \in \mathcal{X}_{t}} \hat{y}^i_t\right) - \mathbb{E}_{x^i_{t} \in \mathcal{X}_{t}} H\left(\hat{y}^i_t\right),
\end{aligned}
\label{Loss_im}
\end{equation}
where $H$ is the same Shannon entropy as used in function Eq.(\ref{H_b}) and Eq.(\ref{H_c}).
Increasing the first term promotes the diversity of predictions across the entire target domain, while decreasing the second one encourages prediction certainty for each target sample.

Overall loss function of the dual-teacher knowledge distillation, denoted as $\mathcal{L}_{dt}$, is defined as follows,
\begin{equation}
\mathcal{L}_{dt} = \mathcal{L}_{kd} + \mathcal{L}_{mix} - \mathcal{L}_{im},
\label{loss1}
\end{equation}
where $\mathcal{L}_{dt}$ is both knowledge guided ($\mathcal{L}_{kd}$) and data driven ($\mathcal{L}_{mix}$ and $\mathcal{L}_{im}$).
\subsubsection{Subnetwork Rectification}
Since the pseudo-labels $\hat{y}$ are inherently noisy, knowledge distillation may become suboptimal if the target model overfits incorrect signals from the teachers. To mitigate this issue, we employ a network-based regularization strategy \cite{DBLP:conf/ijcai/PengDL0023}. Specifically, we optimize the Jensen–Shannon divergence between the outputs of a lightweight subnetwork and those of the full target model, which facilitates adaptation on the target domain. In addition, by enlarging the gradient discrepancy between the subnetwork and the full model, we introduce a controlled perturbation that enforces them to capture complementary knowledge, thereby reducing overfitting to noisy pseudo-labels.

We denote the weights of the full target model as $W_{full}$ and those of the subnetwork as $W_{sub}$. A ratio $\gamma \in (0,1)$ is introduced to construct the subnetwork by selecting the first $\gamma \cdot 100 \%$ of the weights from each layer of the target model. For clarity, we represent the outputs of the target model and the subnetwork as $f_t(x_t;W_{full})$ and $f_t(x_t;W_{sub})$, respectively.
The outputs divergence loss $\mathcal{L}_{od}$ is defined as follows:
\begin{equation}
\mathcal{L}_{od} = D_{JS} \left( f_t\left(x_t; W_{{sub}}\right) \middle\| f_t\left(x_t; W_{{full}}\right) \right),
\label{losssd}
\end{equation}
where $D_{JS}$ denotes the Jensen–Shannon (JS) divergence loss. Compared with directly employing KL divergence loss, the JS divergence loss enables the output distributions of the target model and the subnetwork to align more closely and in a more stable manner \cite{DBLP:conf/ijcai/PengDL0023}.

The adaptation process faces two extremes: early on, the subnetwork differs greatly from the full network and introduces harmful errors, while later it becomes too similar, offering little transferable knowledge. To balance these cases, we define a weighted gradient discrepancy loss as follows, 
\begin{equation}
\label{eq:grad}
 \mathcal{L}_{wg} = (1 + \exp{(-H(f_{t}(x_{t}; W_{sub})))}) cosine(g_{full}, g_{sub}),
\end{equation}
where $g_{full} = \frac{\partial \mathcal{L}_{od}}{\partial W_{full}}$, $g_{sub} = \frac{\partial \mathcal{L}_{od}}{\partial W_{sub}}$, and $cosine(\cdot, \cdot)$ stands for the cosine similarity function.
Minimizing cosine similarity to enforce divergent learning, while the left term adaptively adjusts its strength: uncertain subnetwork predictions reduce the weight, whereas confident ones increase it.

The overall loss function for the subnetwork rectification is composed of the above two loss functions as follows:
\begin{equation}
\mathcal{L}_{sr} = \epsilon \cdot \mathcal{L}_{od} + \zeta \cdot \mathcal{L}_{wg},
\label{loss2}
\end{equation}
where $\mathcal{L}_{od}$ aligns the subnetwork and the full target network in the outputs space, $\mathcal{L}_{wg}$ drives the target model to learn diverse representations, and $\epsilon$, $\zeta$ are trade-off hyperparameters.

\subsubsection{Self-Distillation and Prompt Learning}
As optimization in Stage One progresses, the predictions of the target model, $\hat{y}_t$, become increasingly reliable and can be leveraged to refine pseudo-labels $\hat{y}$. We characterize this process as self-distillation, where the model's own predictions are used to update pseudo-labels that guide its subsequent training. Specifically, we employ an exponential moving average (EMA) strategy \cite{laine2016temporal} to update the pseudo-labels $\hat{y}$ at the end of each epoch, stabilizing updates and mitigating noise:
\begin{equation}
\hat{y} \leftarrow \beta \cdot \hat{y} + (1 - \beta) \cdot \hat{y}_t,
\label{lossmixp}
\end{equation}
where $\beta$ is a hyperparameter set to 0.9. The updated $\hat{y}$, which incorporates the target model's predictive knowledge, is then used in subsequent knowledge distillation via Eq.~(\ref{losskd}), realizing the self-distillation process.

To enhance the adaptability of CLIP to the target domain, the target model's predictions are also leveraged to optimize the learnable textual prompts while keeping the CLIP encoders frozen. This prompt learning preserves CLIP's cross-modal alignment from large-scale pretraining and is significantly more parameter-efficient than full fine-tuning or adapters \cite{gao2024Clip}. The prompts can be updated online, allowing CLIP to iteratively refine its predictions in response to the improving target model. Formally, let $w$ denote the learnable prompts and $x_t$ target samples. The adapted CLIP prediction is
\begin{equation}
\hat{y}_c = CLIP(x_t, w),
\end{equation}
and the prompts $w$ are optimized via a consistency-maximization loss $\mathcal{L}_{cm}$ \cite{zhou2022learning,xiao2024adversarial}:
\begin{equation}
\hat{w} = \arg\min_w \mathcal{L}_{cm},
\label{OptimalPrompts}
\end{equation}
where $\mathcal{L}_{cm} = -\hat{y}_{t} \cdot \hat{y}_{c}$.

{
During Stage One, the fused pseudo-labels $\hat{y}$ are refined via Eq.~(14) at each epoch and carried over to the subsequent epoch. The learnable prompts are updated every five epochs; after each prompt update, the CLIP predictions $\hat{y}_c$ are recomputed and the fused pseudo-labels $\hat{y}$ are regenerated via Eq.~(6), serving as the new initialization for subsequent EMA-based refinement. Thus, adaptive prediction fusion is performed only before Stage-One training and after each periodic prompt update.
}

\subsubsection{Training Objective of Stage One}
The optimal target model is formulated as
\begin{equation}
\hat{f_t} = \arg\min_{f_t} (\mathcal{L}_{dt} + \mathcal{L}_{sr}),
\label{OptimalTargetModel}
\end{equation}
where $\mathcal{L}_{dt}$ and $\mathcal{L}_{sr}$ are defined in Eq.(\ref{loss1}) and Eq.(\ref{loss2}). Since the loss $\mathcal{L}_{cm}$ is exclusively used to optimize the learnable prompts of CLIP, it is not included in Eq.~(\ref{OptimalTargetModel}) which is employed to train the target model.
\subsection{Stage Two} 
{Although Stage One progressively refines the pseudo-labels and ViL prompts, the resulting supervision inevitably remains noisy due to the absence of ground-truth target labels. We therefore employ prototype-based self-training, a technique widely used in domain adaptation \cite{liang2020we}, to further refine the target model. We emphasize that prototype-based self-training itself is not a methodological novelty of DDSR. Unlike conventional SFDA, where adaptation typically starts from a source-trained model, DDSR cannot access or inherit source-model parameters. Therefore, the prototypes in Stage Two are constructed from the target-adapted representations learned through dual-teacher distillation and subnetwork rectification in Stage One. Stage Two thus serves to consolidate the target-domain class structure established in Stage One and further refine the target model.}

Inspired by prior works \cite{caron2018deep,liang2021source,liang2020we}, we first compute class-wise prototypes $\mu$ based on the features extracted by the target model and its predicted class assignments as follows, 
\begin{equation}
\mu_c=\frac{\sum_{i=1}^{n_c} p_i^c q_i}{\sum_{i=1}^{n_c} p_i^c}, 1 \leq c \leq C,
\end{equation}
where $p_i^c$ denotes the probability predicted by the target model that sample $x^i_t$ belongs to class $c$, $n_c$ is the number of samples predicted as class $c$, and $q_i=g_t(x^i_t)$, representing the feature of $x^i_t$ extracted by the feature extractor.

Then, we calculate the cosine distance between each sample ${x}^i_t$ and all the prototypes, and assign the category of the nearest prototype as the self-training pseudo-lable $\bar{y}^i_t$ as follows:
\begin{equation}
\bar{y}^i_t=\arg \min _c D\left(q_i, \mu_c\right),
\label{SelfTrainingLabels}
\end{equation} 
where $D(\cdot, \cdot)$ is the cosine distance function.

Finally, the target model is further self-trained through minimizing the cross-entropy loss between $\bar{y}^i_t$ and $\hat{y}^i_t$:
\begin{equation}
    \mathcal{L}_{self} = \mathbb{E}_{x^i_t \in \mathcal{X}_t}  L_{ent} \left( \bar{y}^i_t, \hat{y}^i_t \right),
    \label{losscen}
\end{equation}
where $L_{ent}$ denotes the cross entropy. The training objective of this stage is expressed as 
\begin{equation}
\hat{f_t} = \arg\min_{f_t} \mathcal{L}_{self}.
\label{SelftrainingObject}
\end{equation}

{Both stages are performed without access to ground-truth target labels. The overall training procedure is summarized in Algorithm~$\ref{alg1}$.}

{\textbf{Discussion on category shifts}.
DDSR is developed for the closed-set BBDA setting with identical source and target label spaces ($\mathcal{Y}^s=\mathcal{Y}^t$). Under category shifts, target-private classes may introduce unreliable source-model predictions, while source-private classes may provide irrelevant source knowledge, potentially propagating erroneous supervision through subsequent adaptation. Handling such label-space mismatches would require additional mechanisms for identifying target-private samples and suppressing irrelevant source categories, which we leave for future work.
}

\begin{algorithm}[t!]
    \caption{Training procedure of the proposed DDSR.}
    \renewcommand{\algorithmicrequire}{\textbf{Input:}}
    \renewcommand{\algorithmicensure}{\textbf{Output:}}
    \renewcommand{\algorithmicindent}{1.5em}
    \label{alg1}
    \begin{algorithmic}[1]
        \REQUIRE Unlabeled target data $D_t$; target model $f_t$ before adaptation; subnetwork ratio $\gamma$; frozen ViL model (CLIP); query access to the black-box source model $f_s$; threshold $\tilde{\Delta}GU$; hyperparameters $\epsilon$, $\zeta$, and $\beta$; total number of epochs $T$; number of Stage One epochs $T_1$.
        \ENSURE Adapted target model $\hat{f}_t$.     
        \STATE Initialize the subnetwork $f_{sub}$ from $f_t$ with the subnetwork ratio $\gamma$.
        \STATE {Obtain the black-box predictions $\hat{y}_b$ and the initial CLIP predictions $\hat{y}_c$ on $D_t$.}
        \STATE {Generate the initial fused pseudo-labels $\hat{y}$ via Eq.~(\ref{quanzhong}).}
        \FOR{epoch $= 1:T$}
            \IF{epoch $\leq T_1$}
                \STATE Calculate $\mathcal{L}_{dt}$ via Eq.~(\ref{loss1}).
                \STATE Calculate $\mathcal{L}_{sr}$ via Eq.~(\ref{loss2}).
                \STATE Update the pseudo-labels $\hat{y}$ via Eq.~(\ref{lossmixp}).
                \STATE Update the target model via Eq.~(\ref{OptimalTargetModel}).                
                \IF{$(\text{epoch} \bmod 5) = 0$}
                   \STATE Update the ViL prompts via Eq.~(\ref{OptimalPrompts}).
                   \STATE {Recompute the CLIP predictions $\hat{y}_{c}$ using the updated prompts.}
                   \STATE {Regenerate the fused pseudo-labels $\hat{y}$ via Eq.~(\ref{quanzhong}).}
                \ENDIF
            \ELSE
                \STATE Generate the pseudo-labels $\bar{y}_t^i$ via Eq.~(\ref{SelfTrainingLabels}).
                \STATE Calculate $\mathcal{L}_{self}$ via Eq.~(\ref{losscen}).
                \STATE Update the target model via Eq.~(\ref{SelftrainingObject}).
            \ENDIF
        \ENDFOR       
        \RETURN $\hat{f}_t$.
    \end{algorithmic}
\end{algorithm}

\section{EXPERIMENTS}
\subsection{Datasets}
We evaluate our approach on three widely used domain adaptation benchmark datasets:
\textbf{Office-31} \cite{saenko2010adapting} contains three domains, namely Amazon (A), Webcam (W), and DSLR (D), with 31 categories and about 4,652 images, exhibiting clear domain shifts;
\textbf{Office-Home} \cite{venkateswara2017deep} includes four domains: Art (Ar), Clipart (Cl), Product (Pr), and Real-World (Rw). It covers 65 categories with over 15,000 images and is challenging due to large inter-domain variations in common office and home objects;
\textbf{VisDA-17} \cite{peng2017visda} is a large-scale synthetic-to-real benchmark with 152,000 synthetic source images and 55,000 real target images across 12 categories. 

\subsection{Compared Methods and Evaluation Metric}
The compared methods are divided into three groups. The first group is UDA methods that have access to source domain data, such as MCD \cite{saito2018maximum}, HMA(CAN) \cite{zhou2023homeomorphism}, DAPL \cite{ge2023domain}, PDA \cite{bai2024prompt}, and FixBi+ \cite{na2025bridging}. The second one contains SFDA methods, where source data is unavailable but the pre-trained source model can be inherited by the target model, including SHOT \cite{liang2020we}, AaD \cite{yang2022attracting}, TPDS \cite{tang2024source}, ProDe-R \cite{Tang2025ProDE}, Co-learn++ \cite{zhang2025source}, and DTKI \cite{zhan2026dual}. The third group covers BBDA methods, where neither source data nor source-model parameters are accessible, and only predictions from the source model are provided, including DINE \cite{liang2022dine}, SEAL \cite{xia2024separation}, RAIN \cite{DBLP:conf/ijcai/PengDL0023}, RFC \cite{zhang2024reviewing}, MLR \cite{li2025leveraging}, BBC \cite{tian2024clip}, AEM \cite{xiao2024adversarial}, and PDLR \cite{tian2025learning}. Classification accuracy on target domains is used as the evaluation metric.

\subsection{Implementation Details}
\textbf{Network architecture.} To ensure fair and reproducible comparisons with state-of-the-art black-box domain adaptation methods, all experiments use the same backbone architectures and initialization strategies as in the baseline works \cite{liang2022dine,xiao2024adversarial}. Specifically, the target model is initialized with ImageNet-pretrained weights independent of the source model: ResNet-50 \cite{he2016deep} is used as the feature extractor for Office-31 and Office-Home, and ResNet-101 for VisDA-17. The original fully connected (FC) layer is replaced with a bottleneck layer followed by batch normalization, and the classifier is implemented as an FC layer with weight normalization. No weights from the black-box source model are inherited, ensuring strict adherence to the BBDA protocol. Following previous works including AEM \cite{xiao2024adversarial} and BBC \cite{tian2024clip}, we utilize the CLIP ViT-B/32 architecture for our ViL model to maintain consistency with state-of-the-art methods. 
By strictly following these settings, the reported performance comparisons are conducted under fully identical experimental conditions.

\textbf{Training protocols and hyperparameters.} The same training protocols as previous works \cite{liang2022dine,xiao2024adversarial,tian2024clip} are adopted. We optimize the network using mini-batch stochastic gradient descent with momentum 0.9 and weight decay $1e^{-3}$. The learning rate is $1e^{-2}$ for all datasets, except $1e^{-3}$ for VisDA-17. The batch size is 64. For hyperparameters in our method, we set $\epsilon=0.6$, $\zeta=0.3$ in Eq.~(\ref{loss2}), $\tilde{\Delta} GU = 0.05$ in Eq.~(\ref{quanzhong}), and the ratio $\gamma=0.84$ for subnetwork construction. For VisDA-17, Stage One and Stage Two are trained for 15 and 10 epochs, respectively; for the other datasets, 25 and 10 epochs are used. All reported results are averaged over three random seeds
. All experiments are implemented in PyTorch and run on a single NVIDIA RTX 4090 GPU. The code will be made publicly available upon publication to facilitate reproducibility.

{
\textbf{Hyperparameter-selection protocol.}
Target labels are never used for supervision or optimization during adaptation. We select common values of $\tilde{\Delta} GU=0.05$ and $\gamma=0.84$ through preliminary sensitivity analyses on representative tasks and fix them for all other tasks without task-specific tuning. Since target-domain accuracy is inspected in these preliminary analyses, this procedure is not strictly label-free model selection. Nevertheless, Section~\ref{Experimental Analysis} shows that DDSR remains stable over broad ranges of both parameters, indicating limited sensitivity to their precise values.
}

\begin{table}[!t]
\centering
\caption{Classification accuracy (\%) on Office-31. ``Source'' denotes applying the pre-trained source model. ``\checkmark'' indicates the use of a ViL model. ``Avg.'' is the average accuracy across tasks. DDSR (Stage One) is the variant that only performs Stage One of the complete framework denoted as DDSR (Full). The results of DDSR are reported as mean accuracy $\pm$ standard deviation over three random seeds, while the results of the compared methods are taken from their original papers. \textbf{Bold} indicates the best results in BBDA, \underline{underline} denotes the second best.}
\begin{adjustbox}{max width=\columnwidth}
\begin{tabular}{c | l | c | c c c c c c | c }
\toprule
\textbf{Type} & \textbf{Method} & \textbf{ViL} & A$\rightarrow$D & A$\rightarrow$W & D$\rightarrow$A & D$\rightarrow$W & W$\rightarrow$A & W$\rightarrow$D & Avg. \\
\midrule
\multirow{1}{*}{-} 
 & Source & \ding{55} & 80.7 & 74.7 & 59.1 & 94.8 & 63.6 & 97.6 & 78.4 \\
\midrule
\multirow{5}{*}{UDA}
 & MCD\cite{saito2018maximum} & \ding{55} & 92.2 & 88.6 & 69.5 & 98.5 & 69.7 & 100.0 & 86.5 \\
 & HMA(CAN)\cite{zhou2023homeomorphism} & \ding{55} & 95.8 & 95.1 & 79.3 & 99.3 & 77.6 & 100.0 & 91.2 \\
 & DAPL\cite{ge2023domain} & \checkmark & 81.7 & 80.3 & 81.2 & 81.8 & 81.0 & 81.3 & 81.2 \\
 & PDA\cite{bai2024prompt} & \checkmark & 91.2 & 92.1 & 83.5 & 98.1 & 82.5 & 99.8 & 91.2 \\
 & FixBi+\cite{na2025bridging} & \ding{55} & 97.6 & 96.7 & 78.7 & 99.3 & 79.6 & 100 & 92.0 \\
\midrule
\multirow{5}{*}{SFDA} 
 & SHOT\cite{liang2020we} & \ding{55} & 94.0 & 90.1 & 74.7 & 98.4 & 74.3 & 99.9 & 88.6 \\
 & AaD\cite{yang2022attracting} & \ding{55} & 96.4 & 92.1 & 99.1 & 100.0 & 75.0 & 76.5 & 89.9 \\
 & TPDS\cite{tang2024source} & \ding{55} & 97.1 & 94.5 & 75.7 & 98.7 & 75.7 & 99.8 & 90.2 \\
 & ProDe-R\cite{Tang2025ProDE} & \ding{55} & 94.4 & 92.1 & 79.8 & 95.6 & 79.0 & 98.6 & 89.9 \\
 & Co-learn++\cite{zhang2025source} & \checkmark & 99.6 & 99.0 & 86.3 & 99.1 & 84.8 & 100 & 94.8 \\
 & DTKI\cite{zhan2026dual} & \checkmark & 98.2 & 95.1 & 83.2 & 98.8 & 83.1 & 99.2 & 92.9 \\
\midrule
\multirow{9}{*}{BBDA}
 & DINE\cite{liang2022dine} & \ding{55} & 91.7 & 87.5 & 72.9 & 96.3 & 73.7 & 98.5 & 86.7 \\
 & SEAL\cite{xia2024separation} & \ding{55} & 95.1 & 88.3 & 77.6 & 96.0 & 76.7 & 99.3 & 88.8 \\
 & RAIN\cite{DBLP:conf/ijcai/PengDL0023} & \ding{55} & 93.8 & 88.8 & 75.5 & 96.8 & 76.7 & \underline{99.5} & 88.5 \\
 & RFC\cite{zhang2024reviewing} & \ding{55} & 94.4 & 93.0 & 76.7 & 95.6 & 77.5 & 98.1 & 89.2 \\
 & MLR\cite{li2025leveraging} & \ding{55} & 94.6 & 92.2 & 78.2 & 96.0 & 79.3 & \textbf{99.6} & 90.0 \\
 & BBC\cite{tian2024clip} & \checkmark & 93.8 & 91.7 & 82.4 & 92.7 & 82.2 & 95.6 & 89.8 \\
 & AEM\cite{xiao2024adversarial} & \checkmark & 95.1 & 94.0 & 81.8 & \underline{98.2} & 82.6 & 99.4 & 91.9 \\
 & PDLR\cite{tian2025learning} & \ding{55} & \underline{96.0} & 91.7 & 76.4 & 98.1 & 79.2 & 98.6 & 90.0 \\
 & DDSR (Stage One) & \checkmark & 95.8$\pm$0.27& \underline{94.8$\pm$0.19} & \underline{83.5$\pm$0.19} & 97.2$\pm$0.26& \underline{83.7$\pm$0.13} & 98.2$\pm$0.20& \underline{92.2$\pm$0.09} \\
 & DDSR (Full) & \checkmark & \textbf{96.8$\pm$0.20} & \textbf{95.7$\pm$0.06} & \textbf{84.5$\pm$0.03} & \textbf{98.8$\pm$0.15} & \textbf{84.4$\pm$0.10} & 98.4$\pm$0.30& \textbf{93.1$\pm$0.08} \\
\bottomrule
\end{tabular}
\end{adjustbox}
\label{office31}
\end{table}

\begin{table*}[htb]
\centering
\caption{Classification accuracy (\%) on Office-Home. DDSR (Stage One) is the variant that only performs Stage One of the complete framework denoted as DDSR (Full). DDSR results are reported as mean accuracy $\pm$ standard deviation over three random seeds, while the results of the compared methods are taken from their original papers. \textbf{Bold} indicates the best results in BBDA, \underline{underline} denotes the second best.}
\begin{adjustbox}{max width=\textwidth}
\begin{tabular}{c | l | c | c c c c c c c c c c c c | c}
\toprule
\textbf{Type} & \textbf{Method} & \textbf{ViL} & Ar$\rightarrow$Cl & Ar$\rightarrow$Pr & Ar$\rightarrow$Rw & Cl$\rightarrow$Ar & Cl$\rightarrow$Pr & Cl$\rightarrow$Rw & Pr$\rightarrow$Ar & Pr$\rightarrow$Cl &  Pr$\rightarrow$Rw & Rw$\rightarrow$Ar & Rw$\rightarrow$Cl & Rw$\rightarrow$Pr & Avg. \\
\midrule
\multirow{1}{*}{-} 
 & Source & \ding{55} & 44.2 & 67.8 & 74.2 & 52.6 & 62.7 & 64.5 & 52.3 & 39.7 & 73.5 & 65.3 & 45.7 & 78.1 & 64.9 \\
\midrule
\multirow{5}{*}{UDA}
 & MCD\cite{saito2018maximum} & \ding{55} & 48.9 & 68.3 & 74.6 & 61.3 & 67.6 & 68.8 & 57.0 & 47.1 & 75.1 & 69.1 & 52.2 & 79.6 & 64.1 \\
 & HMA(CAN)\cite{zhou2023homeomorphism} & \ding{55} & 60.6 & 79.1 & 82.9 & 68.9 & 77.5 & 79.3 & 69.1 & 55.9 & 83.5 & 74.6 & 62.3 & 84.4 & 73.2 \\
 & DAPL\cite{ge2023domain} & \checkmark & 54.1 & 84.3 & 84.4 & 74.4 & 83.7 & 85.0 & 74.5 & 54.6 & 84.8 & 75.2 & 54.7 & 83.8 & 74.5 \\
 & PDA\cite{bai2024prompt} & \checkmark & 55.4 & 85.1 & 85.8 & 75.2 & 85.2 & 85.2 & 74.2 & 55.2 & 85.8 & 74.7 & 55.8 & 86.3 & 75.3 \\
 & FixBi+\cite{na2025bridging} & \ding{55} & 62.9 & 79.0 & 81.3 & 68.5 & 80.8 & 78.2 & 67.0 & 60.8 & 83.1 & 76.4 & 64.4 & 86.7 & 74.1 \\
\midrule
\multirow{5}{*}{SFDA}
 & SHOT\cite{liang2020we} & \ding{55} & 57.1 & 78.1 & 81.5 & 68.0 & 78.2 & 78.1 & 67.4 & 54.9 & 82.2 & 73.3 & 58.8 & 84.3 & 71.8 \\
 & AaD\cite{yang2022attracting} & \ding{55} & 59.3 & 79.3 & 82.1 & 68.9 & 79.8 & 79.5 & 67.2 & 57.4 & 83.1 & 72.1 & 58.5 & 85.4 & 72.7 \\
 & TPDS\cite{tang2024source} & \ding{55} & 59.3 & 80.3 & 82.1 & 70.6 & 79.4 & 80.9 & 69.8 & 56.8 & 82.1 & 74.5 & 61.2 & 85.3 & 73.5 \\
 & ProDe-R\cite{Tang2025ProDE} & \ding{55} & 64.0 & 90.0 & 88.3 & 81.1 & 90.1 & 88.6 & 79.8 & 65.4 & 89.0 & 80.9 & 65.5 & 90.2 & 81.1 \\
 & Co-learn++\cite{zhang2025source} & \checkmark & 80.0 & 91.2 & 91.8 & 83.4 & 92.7 & 91.3 & 83.4 & 78.9 & 92.0 & 85.5 & 80.6 & 94.7 & 87.1 \\
 & DTKI\cite{zhan2026dual} & \checkmark & 72.0 & 91.7 & 90.0 & 82.3 & 91.9 & 90.0 & 82.8 & 71.1 & 90.3 & 82.7 & 71.7 & 91.6 & 84.0 \\
\midrule
\multirow{9}{*}{BBDA}
 & DINE\cite{liang2022dine} & \ding{55} & 54.2 & 77.9 & 81.6 & 65.9 & 77.7 & 79.9 & 64.1 & 50.5 & 82.1 & 71.1 & 58.0 & 84.3 & 70.6 \\
 & SEAL\cite{xia2024separation} & \ding{55} & 58.5 & 81.4 & 84.7 & 71.7 & 80.4 & 82.1 & 72.2 & 54.3 & 86.0 & 76.2 & 60.6 & 86.3 & 74.5 \\
 & RAIN\cite{DBLP:conf/ijcai/PengDL0023} & \ding{55} & 57.0 & 79.7 & 82.8 & 67.9 & 79.5 & 81.2 & 67.7 & 53.2 & 84.6 & 73.3 & 59.6 & 85.6 & 73.0 \\
 & RFC\cite{zhang2024reviewing} & \ding{55} & 57.4 & 80.0 & 82.8 & 67.0 & 80.6 & 80.2 & 68.3 & 57.8 & 82.8 & 72.8 & 59.3 & 85.9 & 72.9 \\
 & MLR\cite{li2025leveraging} & \ding{55} & 57.6 & 80.3 & 82.7 & 68.7 & 78.0 & 79.7 & 66.6 & 57.3 & 81.7 & 74.1 & 60.4 & 86.3 & 72.8 \\
 & BBC\cite{tian2024clip} & \checkmark & 67.5 & 87.4 & 87.1 & 76.4 & 89.0 & 87.3 & 76.7 & 66.7 & 87.3 & 78.1 & 66.2 & 89.3 & 79.9 \\
 & AEM\cite{xiao2024adversarial} & \checkmark & 65.4 & 88.3 & 89.5 & 80.1 & \underline{90.7} & 89.7 & 78.9 & 61.4 & 89.9 & 79.2 & 63.6 & \underline{90.8} & 80.6 \\
 & PDLR\cite{tian2025learning} & \ding{55} & 59.9 & 79.9 & 84.7 & 74.0 & 80.7 & 82.9 & 73.3 & 55.1 & 85.6 & 76.8 & 61.0 & 88.2 & 75.2 \\
 & DDSR (Stage One)& \checkmark & \underline{68.7$\pm$0.10} & \underline{90.1$\pm$0.23} & \underline{90.0$\pm$0.25} & \underline{80.8$\pm$0.15} & 90.0$\pm$0.25 & \underline{89.9$\pm$0.11} & \underline{80.8$\pm$0.20} & \underline{68.3$\pm$0.20} & \underline{90.0$\pm$0.05} & \underline{81.2$\pm$0.25} & \underline{68.6$\pm$0.10} & 90.4$\pm$0.10 & \underline{82.4$\pm$0.07} \\
 & DDSR (Full) & \checkmark & \textbf{70.0$\pm$0.20} & \textbf{91.1$\pm$0.25} & \textbf{90.1$\pm$0.20} & \textbf{81.5$\pm$0.22} & \textbf{91.0$\pm$0.10} & \textbf{90.3$\pm$0.20} & \textbf{81.6$\pm$0.25} & \textbf{69.8$\pm$0.05} & \textbf{90.3$\pm$0.05} & \textbf{81.8$\pm$0.15} & \textbf{70.0$\pm$0.10} & \textbf{91.1$\pm$0.05} & \textbf{83.2$\pm$0.10} \\
\bottomrule
\end{tabular}
\end{adjustbox}
\label{officehome}
\end{table*}

\begin{table*}[htb]
\centering
\caption{Classification accuracy (\%) on VisDA-17. DDSR (Stage One) is the variant that only performs Stage One of the complete framework denoted as DDSR (Full). The results of DDSR are reported as mean accuracy $\pm$ standard deviation over three random seeds, while the results of the compared methods are taken from their original papers. \textbf{Bold} indicates the best results in BBDA, \underline{underline} denotes the second best.}
\begin{adjustbox}{max width=\textwidth}
\begin{tabular}{c | l | c | c c c c c c c c c c c c | c}
\toprule
\textbf{Type} & \textbf{Method} & \textbf{ViL} & plane & bcycl & bus & car & horse & knife & mcycl & person & plant & sktbrd & train & truck & Avg. \\
\midrule
\multirow{1}{*}{-} 
 & Source & \ding{55} & 57.1 & 11.9 & 50.5 & 73.1 & 43.8 & 4.2 & 60.1 & 12.2 & 55.3 & 21.1 & 87.2 & 6.5 & 43.4 \\
\midrule
\multirow{5}{*}{UDA} 
 & MCD\cite{saito2018maximum} & \ding{55} & 87.0 & 60.9 & 83.7 & 64.0 & 88.9 & 79.6 & 84.7 & 76.9 & 88.6 & 40.3 & 83.0 & 25.8 & 71.9 \\
 & HMA(CAN)\cite{zhou2023homeomorphism} & \ding{55} & 97.6 & 88.4 & 84.3 & 76.0 & 98.4 & 97.1 & 91.3 & 81.4 & 97.0 & 96.7 & 88.8 & 60.7 & 88.1 \\
 & DAPL\cite{ge2023domain} & \checkmark & 97.8 & 83.1 & 88.8 & 77.9 & 97.4 & 91.5 & 94.2 & 79.7 & 88.6 & 89.3 & 92.5 & 62.0 & 86.9 \\
 & PDA\cite{bai2024prompt} & \checkmark & 97.2 & 82.3 & 89.4 & 76.0 & 97.4 & 87.5 & 95.8 & 79.6 & 87.2 & 89.0 & 93.3 & 62.1 & 86.4 \\
 & FixBi+\cite{na2025bridging} & \ding{55} & 96.4 & 88.7 & 90.9 & 89.8 & 98.0 & 95.4 & 93.4 & 87.2 & 96.8 & 95.3 & 91.5 & 43.2 & 88.9 \\
\midrule
\multirow{5}{*}{SFDA} 
 & SHOT\cite{liang2020we} & \ding{55} & 94.3 & 88.5 & 80.1 & 57.3 & 93.1 & 94.9 & 80.7 & 80.3 & 91.5 & 89.1 & 86.3 & 58.2 & 82.9 \\
 & AaD\cite{yang2022attracting} & \ding{55} & 97.4 & 90.5 & 80.8 & 76.2 & 97.3 & 96.1 & 89.8 & 82.9 & 95.5 & 93.0 & 92.0 & 64.7 & 88.0 \\
 & TPDS\cite{tang2024source} & \ding{55} & 97.6 & 91.5 & 89.7 & 83.4 & 97.5 & 96.3 & 92.2 & 82.4 & 96.0 & 94.1 & 90.9 & 40.4 & 87.6 \\
 & ProDe-R\cite{Tang2025ProDE} & \ding{55} & 96.6 & 90.3 & 83.9 & 80.2 & 96.1 & 96.9 & 90.3 & 86.4 & 90.8 & 94.0 & 91.3 & 67.0 & 88.7 \\
 & Co-learn++\cite{zhang2025source} & \checkmark & 99.6 & 94.6 & 90.9 & 77.8 & 99.6 & 99.0 & 96.4 & 80.1 & 90.0 & 99.2 & 96.3 & 70.1 & 91.1 \\
 & DTKI\cite{zhan2026dual} & \checkmark & 97.7 & 87.7 & 87.5 & 82.7 & 97.3 & 98.3 & 93.3 & 85.1 & 95.3 & 96.3 & 94.0 & 73.9 & 90.8 \\
\midrule
\multirow{9}{*}{BBDA} 
 & DINE\cite{liang2022dine} & \ding{55} & 95.3 & 85.9 & 80.1 & 53.4 & 93.0 & 37.7 & 80.7 & 79.2 & 86.3 & 89.9 & 85.7 & 60.4 & 77.3 \\
 & SEAL\cite{xia2024separation} & \ding{55} & 97.9 & 92.2 & 88.0 & 73.5 & 97.1 & 96.1 & 92.4 & 85.7 & 93.9 & 95.6 & 91.2 & 66.4 & 89.2 \\
 & RAIN\cite{DBLP:conf/ijcai/PengDL0023} & \ding{55} & 96.6 & 86.8 & 83.0 & 70.9 & 94.5 & 81.8 & 84.2 & 83.6 & 90.9 & 89.5 & 89.4 & 64.0 & 82.7 \\
 & RFC\cite{zhang2024reviewing} & \ding{55} & 95.6 & 89.7 & 87.8 & 75.8 & 96.5 & 96.5 & 90.4 & 82.8 & \textbf{96.0} & 70.0 & 85.7 & 55.1 & 85.2 \\
 & MLR\cite{li2025leveraging} & \ding{55} & 96.5 & 88.5 & 82.6 & 69.4 & 96.0 & 96.3 & 89.3 & 81.8 & 96.0 & 95.7 & 93.1 & 56.2 & 86.8 \\
 & BBC\cite{tian2024clip} & \checkmark & \underline{98.5} & \underline{92.7} & 87.3 & \textbf{78.5} & 98.1 & \textbf{97.6} & 92.1 & 84.7 & 93.3 & 96.5 & \underline{95.1} & \underline{72.8} & \textbf{90.6} \\
 & AEM\cite{xiao2024adversarial} & \checkmark & \textbf{98.6} & 88.1 & \underline{89.7} & 74.8 & 98.0 & 93.9 & 93.0 & \textbf{89.3} & 90.1 & 97.2 & \textbf{95.2} &63.5 & 89.3 \\
 & PDLR\cite{tian2025learning} & \ding{55} & \underline{98.5} & \textbf{93.4} & \textbf{90.6} & \underline{77.7} & \underline{98.3} & 95.7 & \textbf{93.6} & \underline{87.9} & \textbf{96.0} & 94.2 & 93.0 & 62.4 & \underline{90.1} \\
 & DDSR (Stage One) & \checkmark & 98.2$\pm$0.13 & 89.5$\pm$0.20 & 89.3$\pm$0.08 & 75.0$\pm$0.17 & 97.8$\pm$0.11 & \underline{97.1$\pm$0.23} & \textbf{93.6$\pm$0.09} & 83.1$\pm$0.15 & 92.0$\pm$0.07 & \textbf{97.5$\pm$0.19} & 93.4$\pm$0.12 & 69.3$\pm$0.20 & 89.7$\pm$0.10 \\
 & DDSR (Full) & \checkmark & 98.4$\pm$0.14 & 92.2$\pm$0.10 & \underline{89.7$\pm$0.22} & 75.1$\pm$0.18 & \textbf{98.4$\pm$0.06} & \textbf{97.6$\pm$0.11} & \underline{93.2$\pm$0.20} & 83.2$\pm$0.13 & \underline{94.0$\pm$0.08} & \underline{97.3$\pm$0.17} & 94.2$\pm$0.25 & \textbf{73.1$\pm$0.09} & \textbf{90.6$\pm$0.09} \\
\bottomrule
\end{tabular}
\end{adjustbox}
\label{visda}
\end{table*}

\subsection{Experimental Results}
Tables \ref{office31}, \ref{officehome}, and \ref{visda} present results on Office-31, Office-Home, and VisDA, where the best scores in BBDA are shown in \textbf{bold} and the second best are \underline{underlined}. All compared methods are grouped into three categories: UDA, SFDA, and BBDA. The ``Source'' baseline directly applies the pre-trained source model to predict target samples. In the ViL column, ``\checkmark'' and ``\ding{55}'' denote whether a ViL model is used. ``Avg.'' indicates the average accuracy across all tasks in each dataset. Results of all compared methods are taken from the original papers. In addition to the full DDSR framework, we evaluate an ablation variant denoted as DDSR (Stage One), which performs only Stage One of our method without the subsequent prototype-based self-training in Stage Two. This comparison helps quantify the contribution of Stage Two to the overall performance.

\textbf{Office-31.} 
As shown in Table \ref{office31}, DDSR (Full) outperforms the other BBDA methods on all tasks except W$\rightarrow$D, where a slight performance drop is observed. It achieves the highest average accuracy of 93.1\% among the compared BBDA methods, exceeding AEM and BBC by 1.2 and 3.3 percentage points, respectively. Both AEM and BBC are recent BBDA methods that leverage ViLs. Moreover, DDSR (Full), which incorporates both Stage One and Stage Two, consistently outperforms DDSR (Stage One) across all tasks, demonstrating the contribution of Stage Two to the overall adaptation performance. 
{Compared with UDA and SFDA methods that have access to source data or source-model parameters, DDSR does not consistently achieve the best performance but remains competitive with several of these methods despite operating under the more restrictive BBDA setting.}

\textbf{Office-Home.}
As shown in Table \ref{officehome}, DDSR achieves the best performance among the compared BBDA methods on all Office-Home tasks. In particular, on Rw$\rightarrow$Cl, it exceeds the second-best BBDA method, BBC, by 3.8 percentage points. DDSR further achieves the highest average accuracy of 83.2\% among the compared BBDA methods, outperforming AEM and BBC by 2.6 and 3.3 percentage points, respectively.

\textbf{VisDA-17.}
As shown in Table \ref{visda}, DDSR (Full) achieves the highest average accuracy among the compared BBDA methods and ranks first or second among BBDA methods on more than half of the individual tasks. DDSR (Full) also outperforms DDSR (Stage One) on the majority of tasks and achieves higher average accuracy, indicating that Stage Two contributes positively to the overall adaptation performance, although the improvements vary across individual tasks. 

{Overall, DDSR achieves state-of-the-art performance among the compared BBDA methods across the three benchmarks, while several UDA and SFDA methods with stronger source-side access remain competitive or superior. We additionally report standard deviations over three random seeds in Tables~\ref{office31}--\ref{visda}. The consistently small variations (approximately 0.05--0.30 percentage points) indicate stable performance across different random seeds.
}

\begin{figure*}[t!]
    \centering 
    \subfigure[Source (D$\rightarrow$A)]{
        \includegraphics[width=0.22\linewidth]{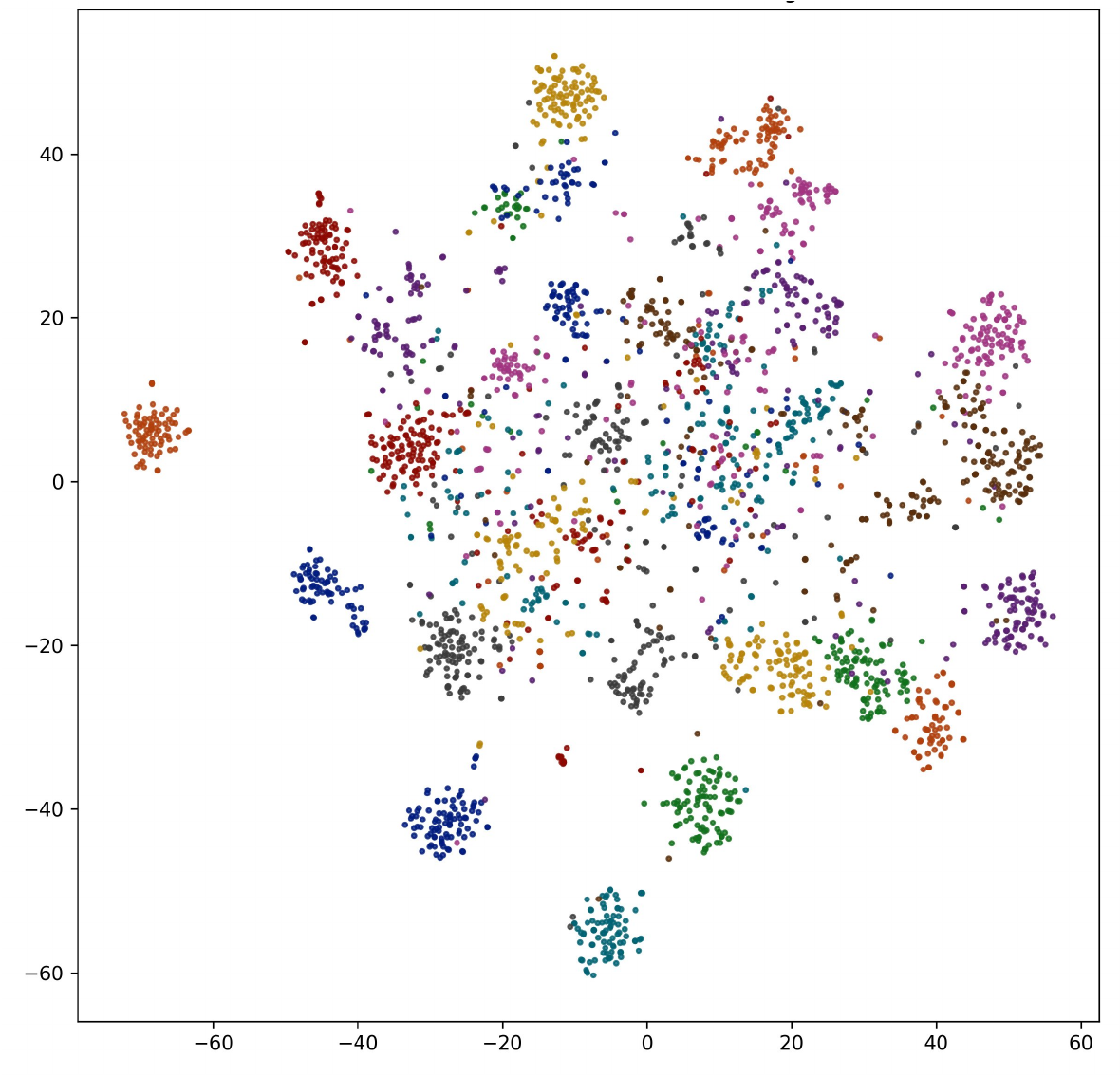} 
    }
    \subfigure[Ours (D$\rightarrow$A)]{
        \includegraphics[width=0.22\linewidth]{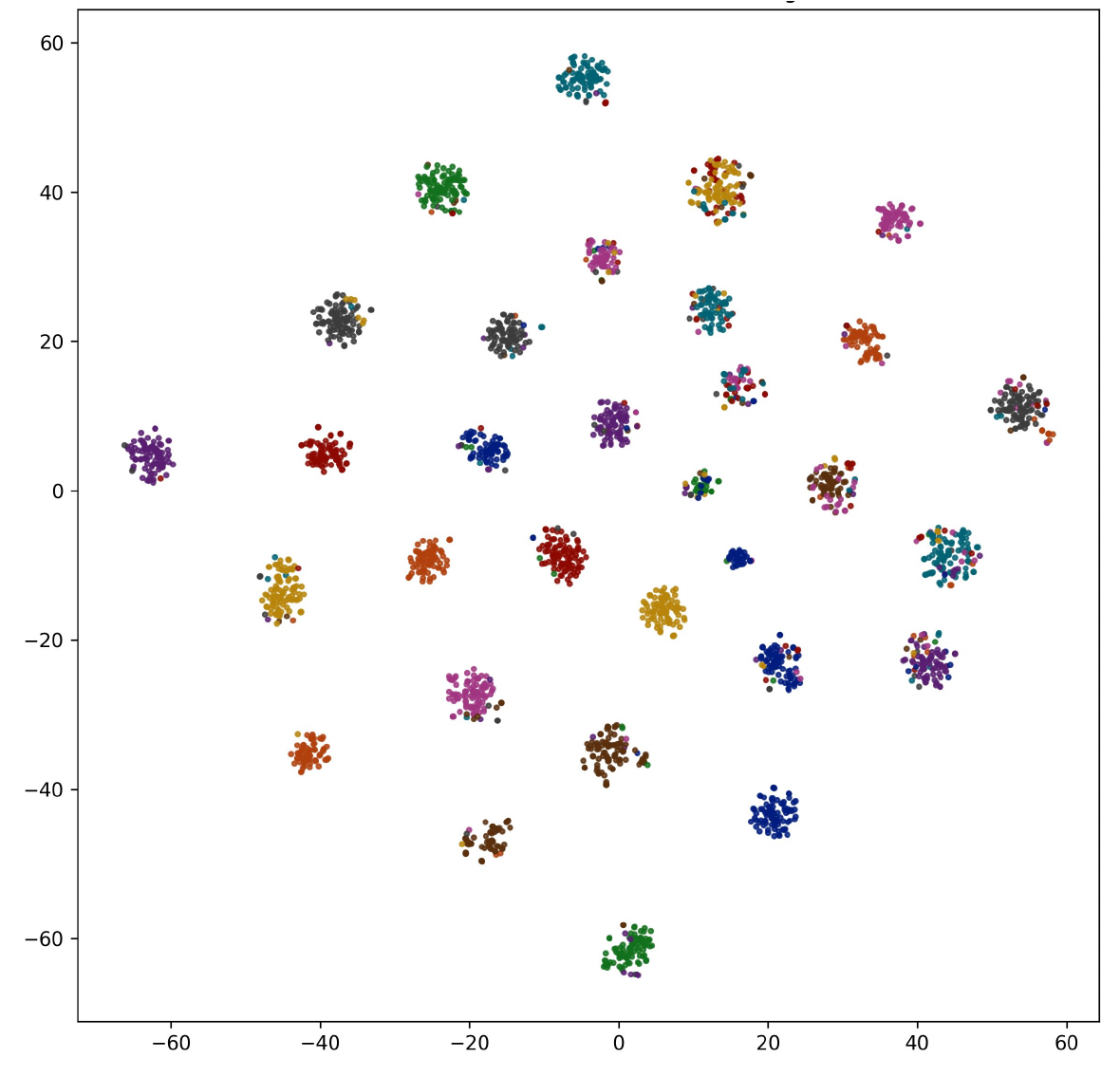}
    }
    \subfigure[Source (W$\rightarrow$A)]{
        \includegraphics[width=0.22\linewidth]{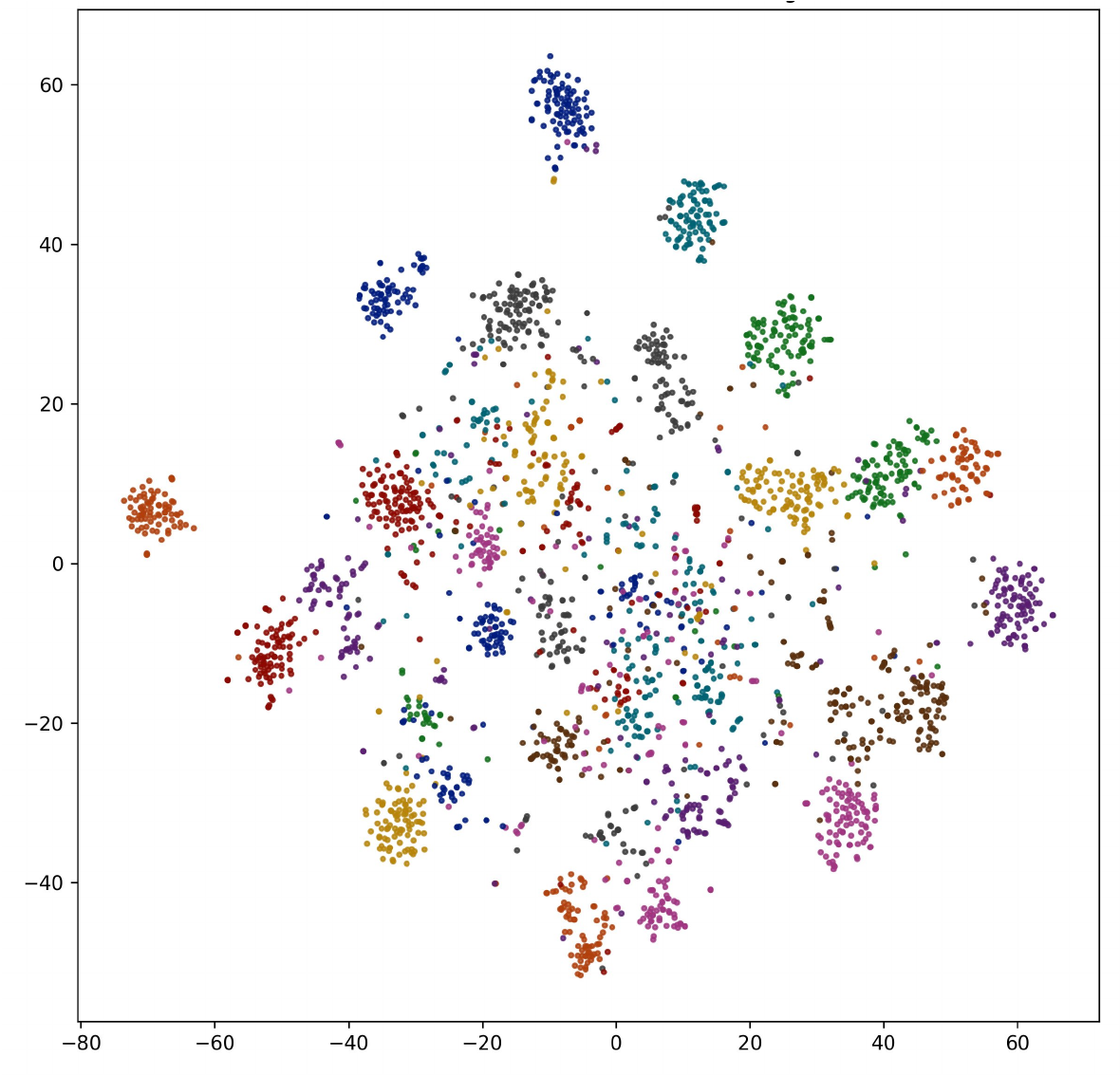}
    }
    \subfigure[Ours (W$\rightarrow$A)]{
        \includegraphics[width=0.22\linewidth]{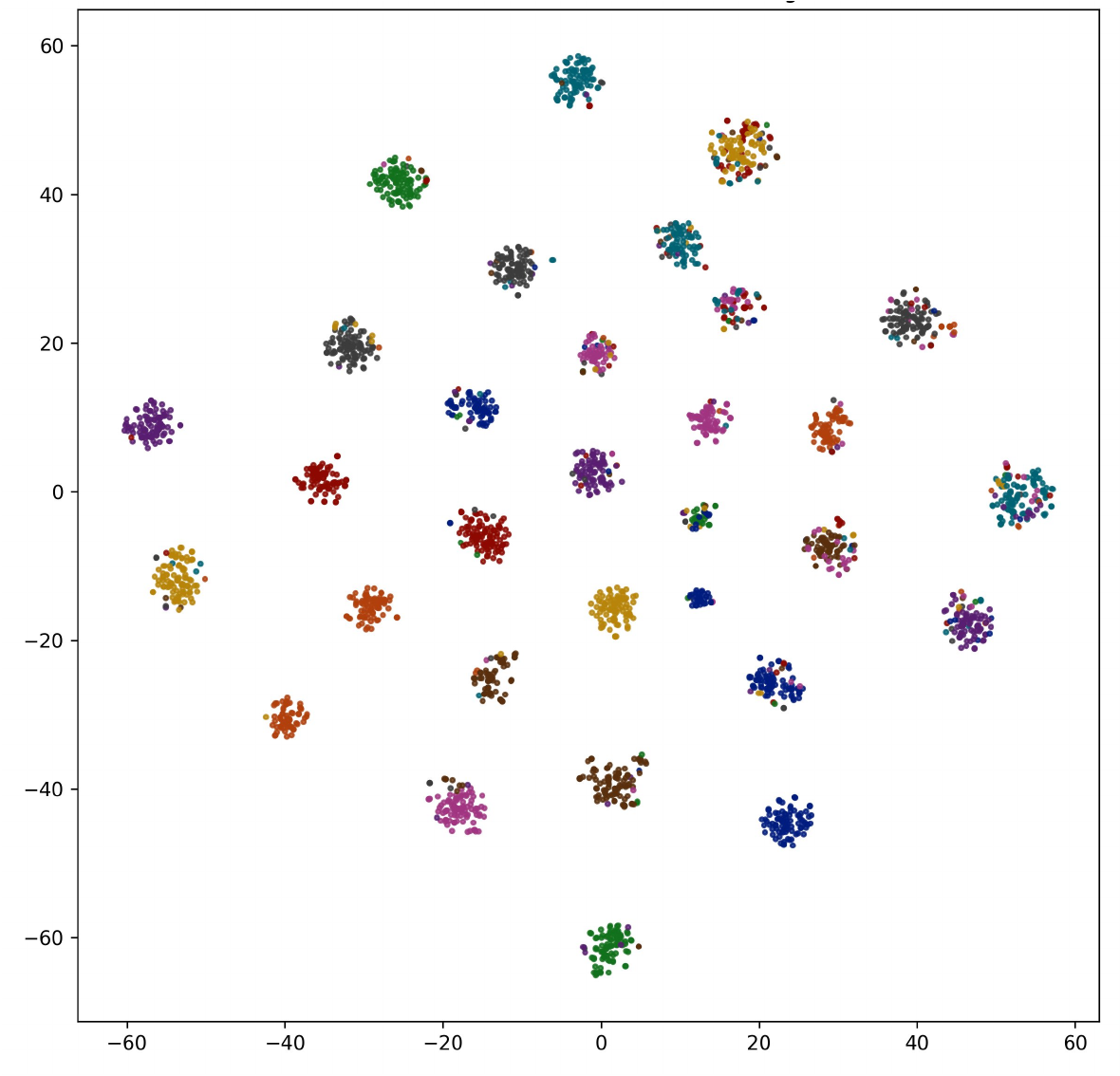}
    }
    \caption{t-SNE visualizations of target features produced by the source model and DDSR on D$\rightarrow$A and W$\rightarrow$A of Office-31. Each point represents a target sample in the feature space, with colors indicating different classes. (Best viewed in color and with magnification.)
    }
    \label{sne}
\end{figure*}

\begin{figure}[t!]
    \centering 
    \subfigure[Ar$\rightarrow$Cl]{
        \includegraphics[width=0.465\linewidth]{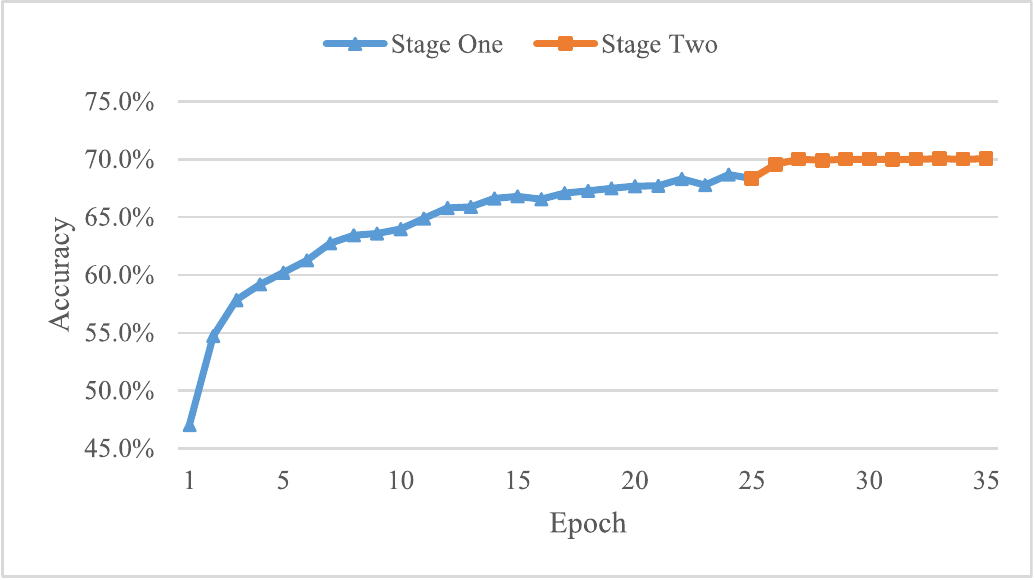} 
        \label{con1}
    }
    \hfill
    \subfigure[D$\rightarrow$A]{
        \includegraphics[width=0.465\linewidth]{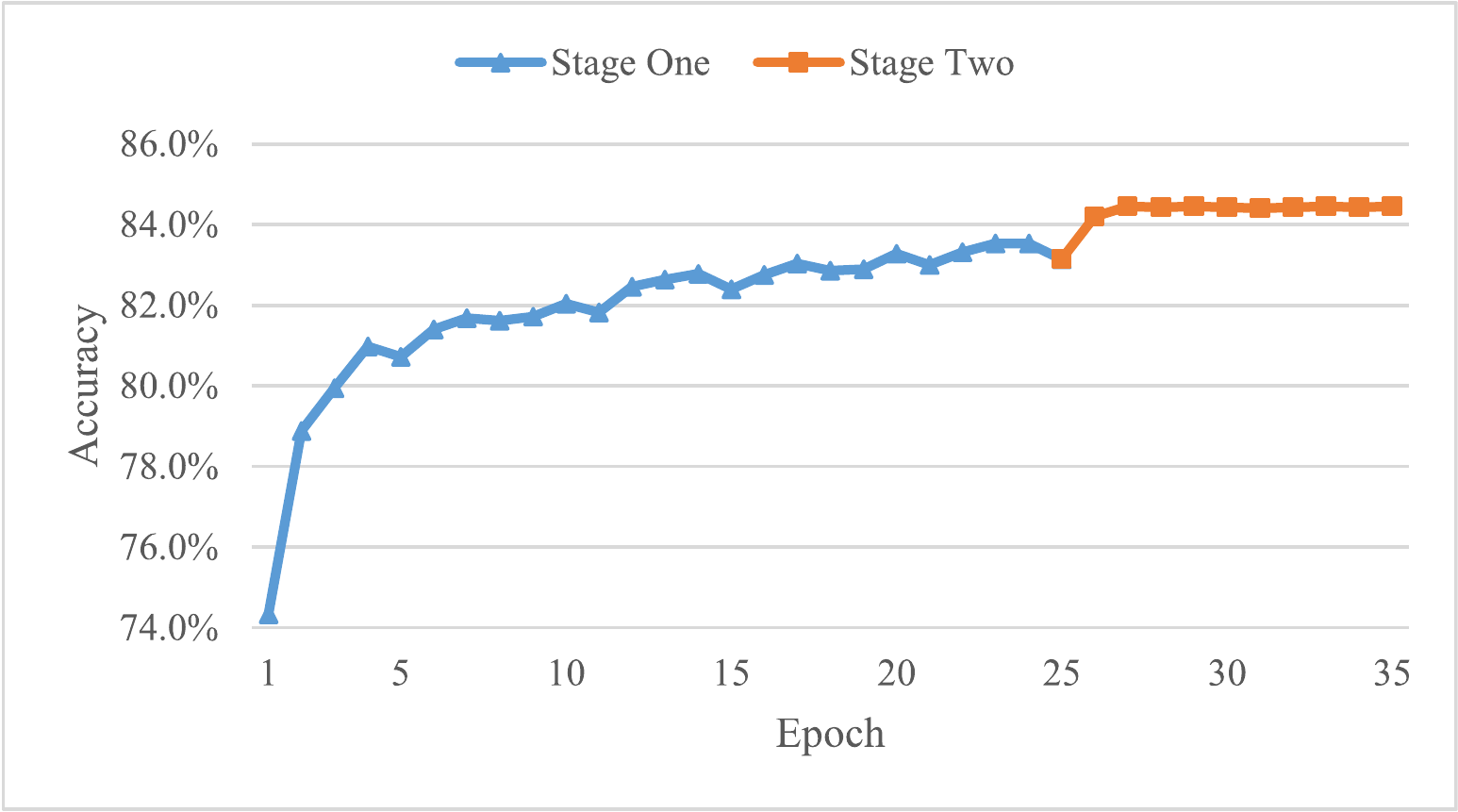}
        \label{con2}
    }
    \caption{Training convergence and stability. Accuracy curves on Ar$\rightarrow$Cl (Office-Home) and D$\rightarrow$A (Office-31) tasks show rapid performance improvement followed by stable convergence.}
    \label{convergence}
\end{figure}

\begin{figure}[t!]
\centering
\includegraphics[width=0.46\textwidth]{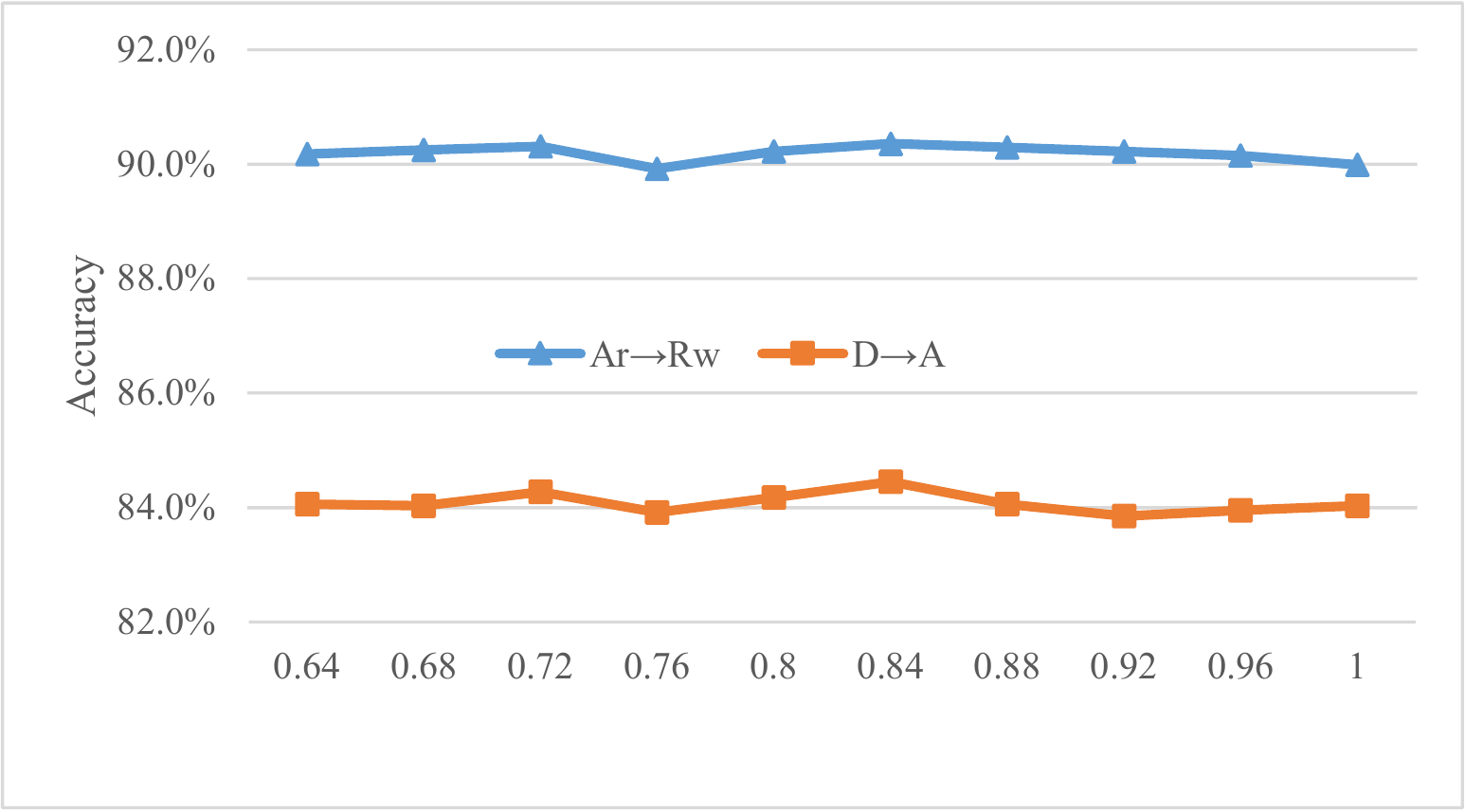}
\caption{{
Sensitivity to the subnetwork ratio $\gamma$ on Ar$\rightarrow$Rw and D$\rightarrow$A.
}}
\label{subrate}
\end{figure}

\begin{figure}[t!]
\centering
\includegraphics[width=0.46\textwidth]{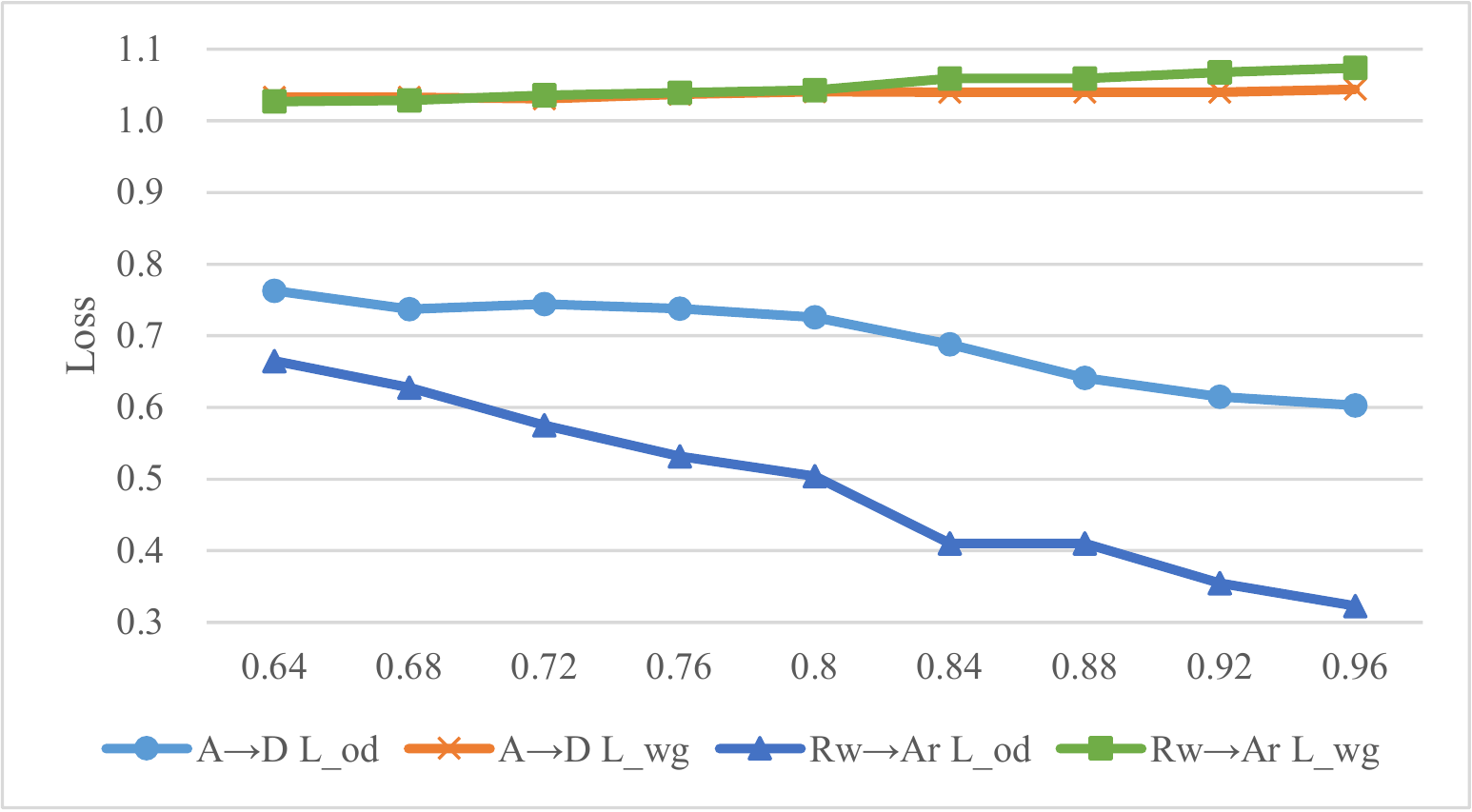}
\caption{{
Values of \(\mathcal{L}_{od}\) and \(\mathcal{L}_{wg}\) under different subnetwork ratios \(\gamma\) on A\(\rightarrow\)D (Office-31) and Rw\(\rightarrow\)Ar (Office-Home). 
}}
\label{ratio_sen}
\end{figure}

\subsection{Experimental Analysis} \label{Experimental Analysis} 
\textbf{Feature visualization.}  
For the adaptation tasks D$\rightarrow$A and W$\rightarrow$A on Office-31, we employ t-SNE \cite{maaten2008visualizing} to visualize the target domain feature space extracted by the source model and our adapted target model. As shown in Fig.\ref{sne}, each point denotes a target sample in the feature space, and different colors indicate distinct  categories. In (a) and (c), the scattered and overlapping distributions produced by the pre-trained source model highlight its inability to cope with domain shifts. In contrast, in (b) and (d), our adapted target model yields well-separated clusters with clear inter-class margins, showing improved feature discrimination and confirming the effectiveness of DDSR in mitigating domain shift.

\textbf{Training convergence and stability.}
As shown in Fig.~\ref{convergence}, for both Ar$\rightarrow$Cl and D$\rightarrow$A tasks, the prediction accuracy of our method increases rapidly in the early training epochs and then gradually converges to a stable level, demonstrating the stability and reliability of the training process.

\textbf{Hyperparameter study.}
\begin{table}[h!]
    \renewcommand{\arraystretch}{1.05} 
    \caption{Accuracy (\%) under different values of the threshold $\tilde{\Delta} GU$ on several tasks of Office-31 and Office-Home. The $\Delta GU$ values for A$\rightarrow$D, A$\rightarrow$W, D$\rightarrow$A, Ar$\rightarrow$Pr, and Cl$\rightarrow$Ar are 0.07, 0.09, 0.04, $-0.03$, and 0, respectively. The best performance is consistently achieved when $\tilde{\Delta} GU = 0.05$.}
    \label{Nth_result}
    \centering
    \begin{tabular}{c c c c c c} 
        \hline
        \textbf{$\tilde{\Delta} GU$} & {A$\rightarrow$D} & {A$\rightarrow$W} & {D$\rightarrow$A} & {Ar$\rightarrow$Pr} & {Cl$\rightarrow$Ar} \\
        \hline
        -0.04 & \textbf{96.8} & \textbf{95.7} & 83.4 & 90.7 & 81.1 \\
        -0.01 & \textbf{96.8} & \textbf{95.7} & 83.4 & \textbf{91.1} & 81.1 \\
        0.02 & \textbf{96.8} & \textbf{95.7} & 83.4 & \textbf{91.1} & \textbf{81.5} \\
        0.05 & \textbf{96.8} & \textbf{95.7} & \textbf{84.5} & \textbf{91.1} & \textbf{81.5} \\
        0.08 & 91.3 & \textbf{95.7} & \textbf{84.5} & \textbf{91.1} & \textbf{81.5} \\
        0.11 & 91.3 & 91.2 & \textbf{84.5} & \textbf{91.1} & \textbf{81.5} \\
        \hline
    \end{tabular}
\end{table}
{Following the protocol described in implementation details, we analyze $\tilde{\Delta} GU$, $\gamma$, $\epsilon$, and $\zeta$ on representative tasks while adopting a common configuration across all adaptation tasks. The EMA coefficient $\beta$ is fixed to 0.9 following common practice~\cite{laine2016temporal} and is not further varied. For each sensitivity analysis, all other parameters are fixed.}

According to Eq.~(\ref{quanzhong}), the selection of different fusion strategies for $\hat{y}_{b}$ and $\hat{y}_{c}$ is determined by the relationship between the threshold $\tilde{\Delta} GU$ and $\Delta GU$. As long as this relationship remains unchanged, the fusion strategy also remains fixed, leading to stable model performance. To determine an appropriate value for $\tilde{\Delta} GU$, we assess its impact on several representive tasks in Office-31 and Office-Home by varying its values. The $\Delta GU$ values for A$\rightarrow$D, A$\rightarrow$W, D$\rightarrow$A, Ar$\rightarrow$Pr, and Cl$\rightarrow$Ar are 0.07, 0.09, 0.04, $-0.03$, and 0, respectively. As shown in Table~\ref{Nth_result}, the accuracy consistently reaches its peak when $\tilde{\Delta} GU$ is set to 0.05, supporting the appropriateness of this choice. For example, in the A$\rightarrow$D task, the $\Delta GU$ for the DSLR (D) domain is 0.07. When $\tilde{\Delta} GU$ is set to 0.08 or 0.11, the condition $\Delta GU < \tilde{\Delta} GU$ holds, and the accuracy remains at 91.3\%. In contrast, when $\tilde{\Delta} GU \leq 0.05$, the condition $\Delta GU > \tilde{\Delta} GU$ is satisfied, triggering the alternative fusion strategy and resulting in an accuracy of 96.8\%.

{
The step-like performance changes on some tasks are expected because $\tilde{\Delta}GU$ serves as a branch-selection threshold in Eq.~(6), rather than a continuous fusion coefficient. Once it crosses the task-specific $\Delta GU$, the fusion rule switches between the two branches, potentially causing a discrete performance change. Importantly, $\tilde{\Delta}GU=0.05$ achieves the best performance (including ties) across all evaluated tasks, supporting a common threshold without task-specific tuning.
Therefore, the performance change around the threshold reflects the intended switch between the two fusion regimes rather than sensitivity within a single fusion rule; importantly, the regime selected according to the actual $\Delta GU$ consistently corresponds to the better-performing branch.
}

{To analyze the subnetwork ratio $\gamma$, we vary it from 0.64 to 0.96. As shown in Fig.~~\ref{subrate}, performance remains stable across this range, with $\gamma=0.84$ achieving the best results. Fig.~~\ref{ratio_sen} further shows that $\mathcal{L}_{od}$ generally decreases as $\gamma$ increases because a wider subnetwork more closely approximates the full network, whereas $\mathcal{L}_{wg}$ remains relatively stable. These results indicate that the proposed subnetwork rectification is robust to the precise choice of $\gamma$.}

\begin{figure}[t!]
\centering
\includegraphics[width=0.48\textwidth]{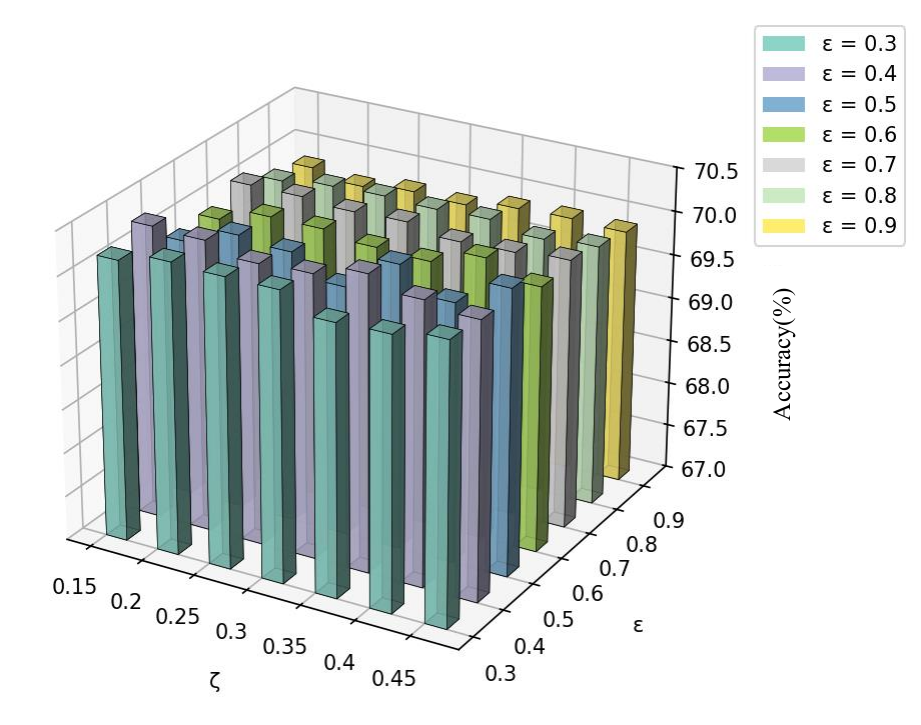}
\caption{Sensitivity to $\epsilon$ and $\zeta$ on Ar$\rightarrow$Cl of Office-Home.
}
\label{losszeta}
\end{figure}

To identify the optimal hyperparameter combination of $\epsilon$ and $\zeta$ in Eq.~(\ref{loss2}), we conducted a sensitivity analysis on the Ar$\rightarrow$Cl task of the Office-Home dataset. As shown in Fig.~\ref{losszeta}, variations in these hyperparameters have a relatively minor impact on the accuracy, indicating that our method maintains stable performance across a broad range of parameter values and thus requires limited hyperparameter tuning.

\begin{figure}[t!]
\centering
\includegraphics[width=0.48\textwidth]{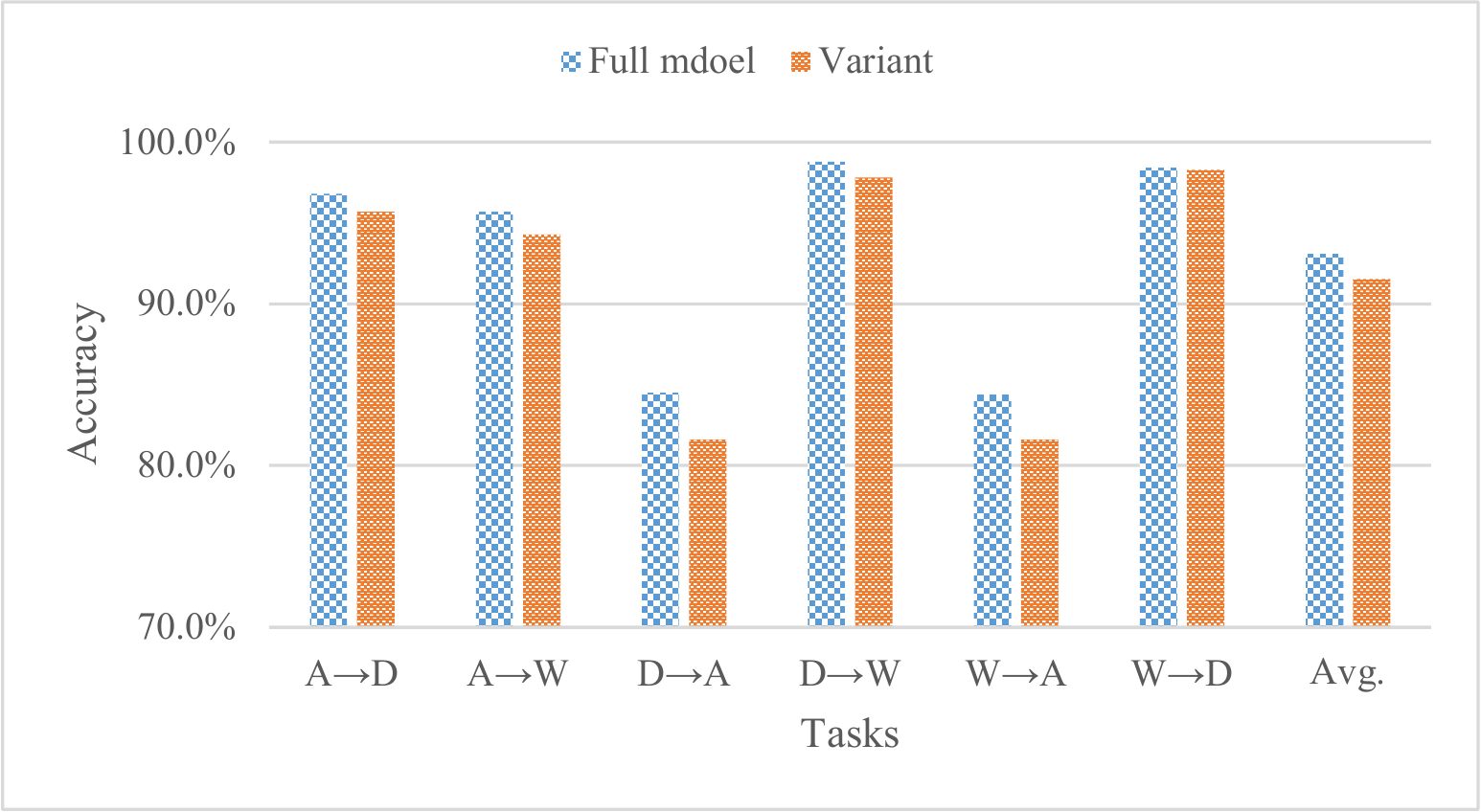}
\caption{Comparison of adaptive prediction fusion and fixed averaging on Office-31.
}
\label{apf}
\end{figure}

\begin{table}[t]
\centering
\caption{{Ablation of adaptive prediction fusion on Office-Home and VisDA-17.}}
\label{fusion_ablation}
\begin{adjustbox}{max width=\columnwidth}
\begin{tabular}{c | c c c c | c}
\toprule
& \multicolumn{4}{c|}{\textbf{Office-Home}} & \textbf{VisDA-17} \\
\textbf{Fusion Strategy} 
& Pr$\rightarrow$Ar 
& Cl$\rightarrow$Pr 
& Cl$\rightarrow$Rw 
& Pr$\rightarrow$Cl 
& S$\rightarrow$R \\
\midrule
Fixed averaging 
& 80.1 & 89.9 & 89.0 & 68.6 & 85.6 \\
Adaptive fusion 
& \textbf{81.6} & \textbf{91.0} & \textbf{90.3} & \textbf{69.8} & \textbf{90.6} \\
\bottomrule
\end{tabular}
\end{adjustbox}
\end{table}

\begin{figure}[t!]
\centering
\includegraphics[width=0.48\textwidth]{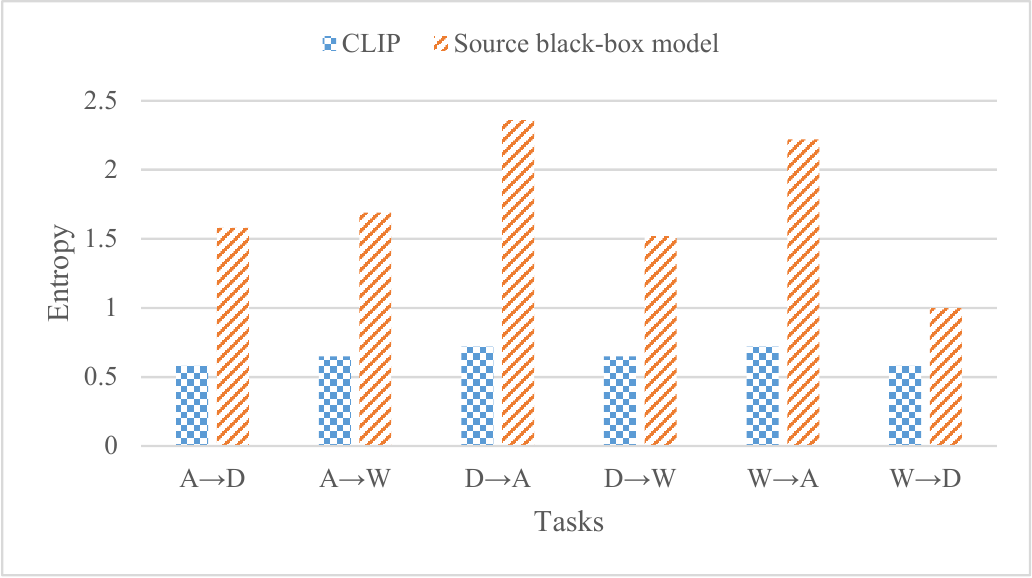}
\caption{Average prediction entropy (IU) of CLIP and the black-box source model on Office-31.}
\label{clip_src_entropy}
\end{figure}

\begin{figure}[t!]
\centering
\includegraphics[width=0.48\textwidth]{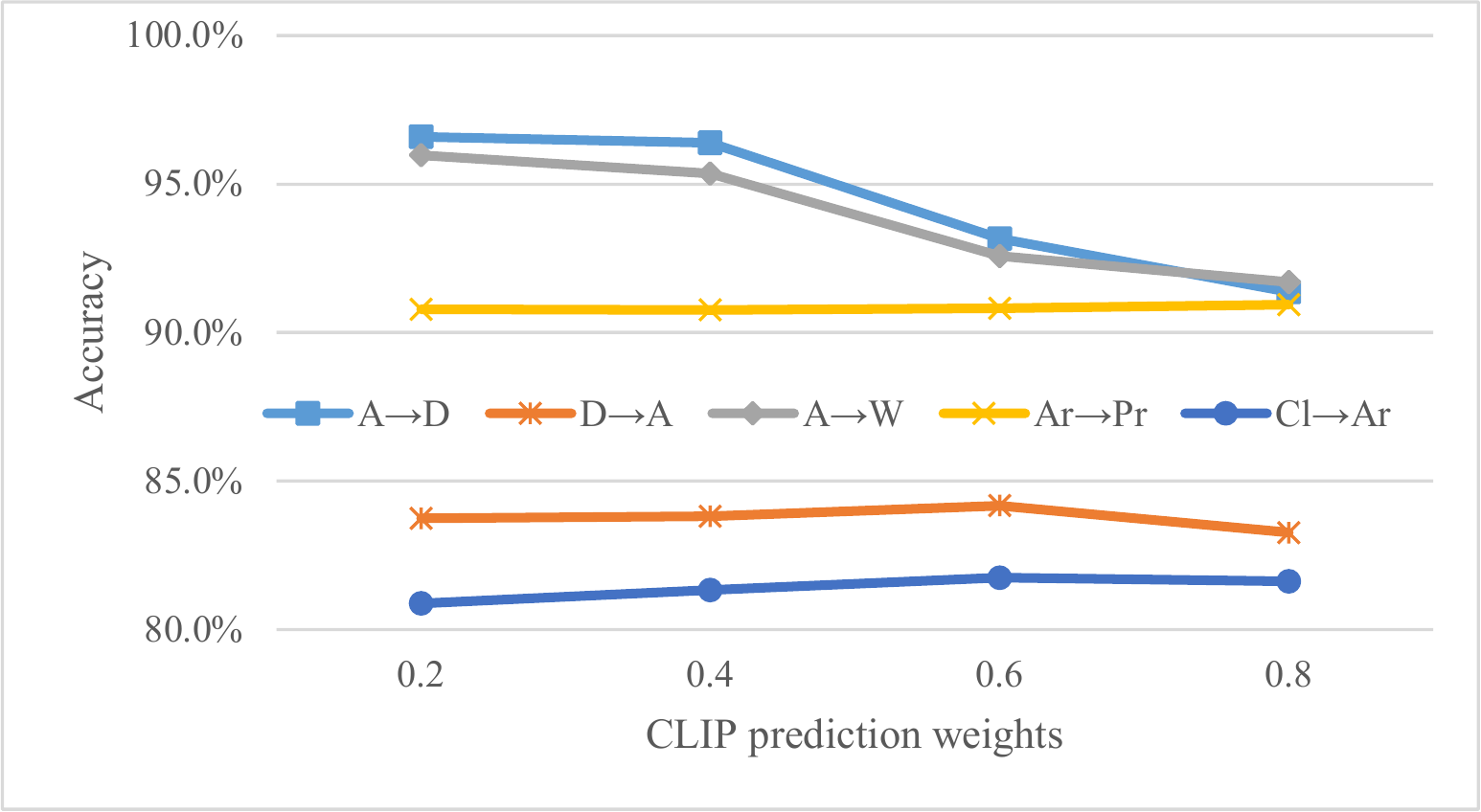}
\caption{Accuracy (\%) under different CLIP prediction weights on representative tasks with different $\Delta GU$ values.}
\label{clip_alpha_acc}
\end{figure}

{
We evaluate the effectiveness of adaptive prediction fusion by comparing it with a variant that replaces Eq.~(\ref{quanzhong}) with fixed averaging of the two teachers. As shown in Fig.~\ref{apf}, adaptive fusion consistently outperforms fixed averaging across all Office-31 tasks, with an average improvement of 1.5 percentage points. To verify its generalizability, we extend this comparison to representative tasks from Office-Home and VisDA-17, as reported in Table~\ref{fusion_ablation}. Adaptive fusion consistently improves performance on all five additional tasks, yielding gains of 1.1--1.5 percentage points on Office-Home and a larger gain of 5.0 percentage points on VisDA-17. These consistent improvements across all three benchmarks demonstrate that adaptively balancing the two complementary teachers is more effective than assigning them fixed equal weights.
}

{
We further examine the confidence--diversity trade-off underlying Eq.~(\ref{quanzhong}). As shown in Fig.~\ref{clip_src_entropy}, CLIP consistently exhibits lower IU than the black-box source model on Office-31, resulting in $\alpha<0.5$. However, high instance-level confidence does not necessarily imply diverse dataset-level predictions, motivating $\Delta GU$ as a complementary signal. Fig.~\ref{clip_alpha_acc} further shows that tasks with larger $\Delta GU$ generally favor smaller CLIP weights, whereas those with smaller $\Delta GU$ favor larger CLIP weights. Together with Table~\ref{Nth_result}, these results empirically support the regime-dependent weighting strategy.}

\begin{table}[t]
\centering
\caption{{Classification accuracy (\%) under the original and artificially imbalanced target-domain class distributions on representative tasks from Office-31 and Office-Home.}}
\begin{adjustbox}{max width=\columnwidth}
\begin{tabular}{c|cc|ccc}
\toprule
& \multicolumn{2}{c|}{\textbf{Office-31}}
& \multicolumn{3}{c}{\textbf{Office-Home}} \\
\textbf{Setting}
& D$\rightarrow$A
& W$\rightarrow$A
& Cl$\rightarrow$Ar
& Pr$\rightarrow$Ar
& Rw$\rightarrow$Ar \\
\midrule
Original
& 84.5 & 84.4 & 81.5 & 81.6 & 81.8 \\
Imbalanced
& 85.0 & 85.2 & 69.2 & 69.1 & 71.1 \\
\bottomrule
\end{tabular}
\end{adjustbox}
\label{imbalance}
\end{table}

\begin{table}[t]
\centering
\caption{{Ablation study of the loss components on representative tasks from Office-31, Office-Home, and VisDA-17. A ``$\checkmark$'' indicates that the corresponding loss is included, while ``\ding{55}'' indicates that it is removed.}}
\begin{adjustbox}{max width=\columnwidth}
\begin{tabular}{c c c c c | c c c c c}
\toprule
\textbf{$\mathcal{L}_{kd}$}
& \textbf{$\mathcal{L}_{mix}$}
& \textbf{$\mathcal{L}_{im}$}
& \textbf{$\mathcal{L}_{sr}$}
& \textbf{$\mathcal{L}_{self}$}
& D$\rightarrow$A
& W$\rightarrow$A
& Pr$\rightarrow$Cl
& Rw$\rightarrow$Ar
& S$\rightarrow$R \\
\midrule
\checkmark & \ding{55} & \ding{55} & \ding{55} & \ding{55}
& 79.1 & 79.3 & 64.3 & 75.8 & 82.6 \\
\checkmark & \ding{55} & \checkmark & \checkmark & \checkmark
& 83.6 & 83.6 & 68.0 & 79.5 & 86.4 \\
\checkmark & \checkmark & \ding{55} & \checkmark & \checkmark
& 80.3 & 80.4 & 67.1 & 79.5 & 84.9 \\
\checkmark & \checkmark & \checkmark & \ding{55} & \checkmark
& 83.8 & 83.9 & 68.6 & 79.1 & 86.8 \\
\checkmark & \checkmark & \checkmark & \checkmark & \ding{55}
& 83.5 & 83.7 & 67.2 & 79.1 & 85.7 \\
\checkmark & \checkmark & \checkmark & \checkmark & \checkmark
& \textbf{84.5} & \textbf{84.4} & \textbf{69.8} & \textbf{81.8} & \textbf{90.6} \\
\bottomrule
\end{tabular}
\end{adjustbox}
\label{ablationloss}
\end{table}

\begin{table}[t]
\centering
\caption{{Ablation study of EMA-based self-distillation and prompt learning on representative tasks from Office-31, Office-Home, and VisDA-17. All other components of DDSR are kept unchanged.}}
\begin{adjustbox}{max width=\columnwidth}
\begin{tabular}{c|cc|cc|c}
\toprule
\textbf{Variant}
& \multicolumn{2}{c|}{\textbf{Office-31}}
& \multicolumn{2}{c|}{\textbf{Office-Home}}
& \textbf{VisDA-17} \\
& D$\rightarrow$A
& W$\rightarrow$A
& Ar$\rightarrow$Rw
& Cl$\rightarrow$Rw
& S$\rightarrow$R \\
\midrule
w/o EMA-based self-distillation
& 84.2 & 84.2 & 89.8 & 90.0 & 86.8 \\
w/o prompt learning
& 82.4 & 82.5 & 87.7 & 88.1 & 86.1 \\
DDSR
& \textbf{84.5} & \textbf{84.4} & \textbf{90.1} & \textbf{90.3} & \textbf{90.6} \\
\bottomrule
\end{tabular}
\end{adjustbox}
\label{ablationrefinement}
\end{table}

{
\textbf{Robustness to target class imbalance.}
Both GU and the information-maximization objective ($\mathcal{L}_{im}$) encourage diverse marginal predictions, which may conflict with a naturally imbalanced target label prior. We therefore evaluate DDSR under controlled target class imbalance. For Office-31, the first and last 10 classes are subsampled to 30 target samples per class; for Office-Home, the first 50 classes are subsampled to 10 samples per class. The classes are selected solely by their fixed ordering, independent of model predictions or performance. All other settings remain unchanged.
}

{
As shown in Table~\ref{imbalance}, DDSR remains stable on Office-31 but exhibits more substantial degradation on Office-Home under severe imbalance. This reveals a limitation of the current formulation: GU and information maximization do not explicitly distinguish prediction collapse from a naturally low-entropy target label prior. Explicitly estimating or accounting for the target label prior is therefore a promising direction for handling severe class imbalance.
}

\textbf{Ablation study}\label{Ablation}.
{To comprehensively evaluate the contribution of individual components, we conduct ablation studies on representative tasks covering all three benchmarks.}

{
We first investigate the contribution of each loss component, as summarized in Table~VI. Since $\mathcal{L}_{kd}$ serves as the core distillation objective, it is retained in all configurations. Removing any of $\mathcal{L}_{mix}$, $\mathcal{L}_{im}$, $\mathcal{L}_{sr}$, or $\mathcal{L}_{self}$ consistently degrades the overall performance, confirming their complementary contributions. Among them, removing $\mathcal{L}_{im}$ generally causes the largest degradation, while $\mathcal{L}_{sr}$ and $\mathcal{L}_{self}$ provide consistent improvements across the evaluated benchmarks.}

{
The different magnitudes of the improvements from $\mathcal{L}_{im}$ and $\mathcal{L}_{sr}$ reflect their distinct roles. $\mathcal{L}_{im}$ provides a strong prediction-level constraint by promoting confident instance predictions and diverse marginal predictions, whereas the proposed $\mathcal{L}_{sr}$ specifically targets robustness to noisy pseudo-label supervision through subnetwork regularization. Although the gain from $\mathcal{L}_{sr}$ is relatively modest on Office-31, it becomes more pronounced on Office-Home and VisDA-17, demonstrating its complementary model-level regularization effect.}

{
In addition to the loss-function ablations, we further ablate EMA-based self-distillation and prompt learning, which progressively refine the supervision during Stage One. As shown in Table~\ref{ablationrefinement}, removing either component consistently degrades performance. EMA-based refinement yields modest gains on Office-31 and Office-Home but a larger improvement on VisDA-17, while prompt learning provides more consistent gains across all three benchmarks. These results confirm their complementary roles in incorporating emerging target-domain knowledge and adapting the initially task-agnostic ViL supervision, respectively.}

\begin{table}[t]
\centering
\caption{{Target-domain classification accuracy (\%) with different black-box source model architectures. Source (Res/ViT) denotes the black-box source model using ResNet/ViT-B/16. DDSR (Res/ViT) denote DDSR using ResNet/ViT-B/16 as the black-box source model, while the target model is fixed as ResNet.}}
\begin{adjustbox}{max width=\textwidth}
\begin{tabular}{c|ccc}
\toprule
\textbf{Method} & \textbf{Office-31} & \textbf{Office-Home} & \textbf{VisDA-17} \\
\midrule
Source (Res) & 78.4 & 60.1 & 40.3 \\
Source (ViT) & 82.8 & 66.0 & 57.1 \\
DDSR (Res)   & 93.1 & 83.2 & 90.6 \\
DDSR (ViT)   & 91.5 & 83.4 & 88.5 \\
\bottomrule
\end{tabular}
\end{adjustbox}
\label{vit}
\end{table}

{\textbf{Impact of black-box source model architecture}.}

{
To evaluate architectural flexibility, we replace the ResNet-based black-box source model with ViT-B/16 while keeping the ResNet target model unchanged. As shown in Table~\ref{vit}, DDSR maintains comparable performance under this architecture-heterogeneous setting, with changes of $-1.6$, $+0.2$, and $-2.1$ percentage points on Office-31, Office-Home, and VisDA-17, respectively. Interestingly, although Source (ViT) is consistently stronger than Source (Res), DDSR (ViT) does not consistently outperform DDSR (Res), possibly because the ViT source model shares more similar architectural characteristics with CLIP's ViT encoder, reducing the complementarity between the two teachers. These results demonstrate that DDSR does not require the target architecture to match the black-box source model.}

\textbf{Complexity analysis.}
Pseudo-label generation via CLIP relies on cross-modal similarity computations using frozen, pre-trained encoders. As these computations are independent of the optimization of the target model, they are excluded from the method-specific complexity analysis. 
{
Adaptive prediction fusion incurs $\mathcal{O}(n_tC)$ complexity and is performed only before Stage-One training and after each prompt update every five epochs. Stage One is dominated by forward and backward passes of the target network and subnetwork, scaling as $\mathcal{O}(n_td)$, while dual-teacher distillation and periodic prompt updates incur $\mathcal{O}(n_tC)$. Subnetwork rectification additionally involves gradient comparison linear in the number of parameters $P$. Stage Two requires feature extraction, prototype computation, and distance evaluation, scaling as $\mathcal{O}(n_td)$. Therefore, DDSR scales linearly with the number of target samples and avoids the $\mathcal{O}(n_t^2d)$ complexity of repeated full-dataset neighbor search.}

\begin{table}[t]
\centering
\caption{{Training runtime (seconds/epoch) on a single NVIDIA RTX 4090 GPU. ``w/o Subnetwork'' denotes training without subnetwork rectification, ``Regular'' denotes a regular DDSR training epoch, and ``Prompt update'' denotes an epoch involving CLIP prompt updating.}}
\begin{adjustbox}{max width=\columnwidth}
\begin{tabular}{c|ccc|cccc|c}
\toprule
& \multicolumn{3}{c|}{\textbf{Office-31}}
& \multicolumn{4}{c|}{\textbf{Office-Home}}
& \textbf{VisDA-17} \\
\textbf{Setting}
& D$\rightarrow$A
& A$\rightarrow$D
& A$\rightarrow$W
& Cl$\rightarrow$Ar
& Ar$\rightarrow$Cl
& Ar$\rightarrow$Pr
& Ar$\rightarrow$Rw
& S$\rightarrow$R \\
\midrule
w/o Subnetwork
& 41 & 9 & 12 & 32 & 57 & 58 & 57 & 1210 \\
Regular
& 45 & 9 & 13 & 34 & 62 & 61 & 60 & 1292 \\
Prompt update
& 64 & 24 & 28 & 51 & 82 & 82 & 80 & 1626 \\
\bottomrule
\end{tabular}
\end{adjustbox}
\label{runtime}
\end{table}

{We further evaluate training efficiency on a single NVIDIA RTX 4090 GPU, with the results reported in Table~\ref{runtime}. Subnetwork rectification introduces only modest overhead, increasing the runtime of a regular epoch by no more than approximately 10\% across the evaluated tasks (6.8\% on VisDA-17). Prompt-update epochs are more expensive due to prompt optimization, but occur only once every five epochs, limiting their contribution to the overall training cost.}

\section{CONCLUSION}
In this paper, we propose a practical framework for the challenging black-box domain adaptation. Rather than solely relying on the zero-shot capability of ViLs to generate pseudo-labels, our method introduces an adaptive prediction fusion mechanism that dynamically integrates semantic knowledge from the ViL model and task-specific knowledge from the black-box source model. 
The framework consists of two stages. In Stage One, pseudo-labels produced by adaptive prediction fusion guide the training of the target model, with an auxiliary subnetwork employed to mitigate overfitting. In Stage Two, the target model is further refined through self-training.
{Extensive experiments on three benchmark datasets demonstrate that DDSR achieves state-of-the-art performance among existing BBDA methods while remaining competitive with several UDA and SFDA approaches that access stronger source-side information. Nevertheless, a performance gap remains under the more restrictive BBDA setting, and the current framework assumes identical source and target label spaces. Extending DDSR to category-shift scenarios constitutes an important direction for future work.}

\bibliographystyle{IEEEtran}
\bibliography{ref}

\vspace{15 mm}
\begin{IEEEbiography}[{\includegraphics[width=1.0in]{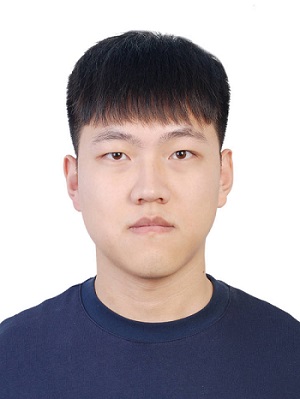}}]{Zhe Zhang}
received the B.S. degree from North China Institute of Aerospace Engineering, Langfang, China, in 2024. He is currently working toward the M.S. degree with the School of Computer Science and Engineering, Tianjin University of Technology, Tianjin, China. His research interests include computer vision and machine learning.
\end{IEEEbiography}

\vspace{-5 mm}
\begin{IEEEbiography}[{\includegraphics[width=1.0in]{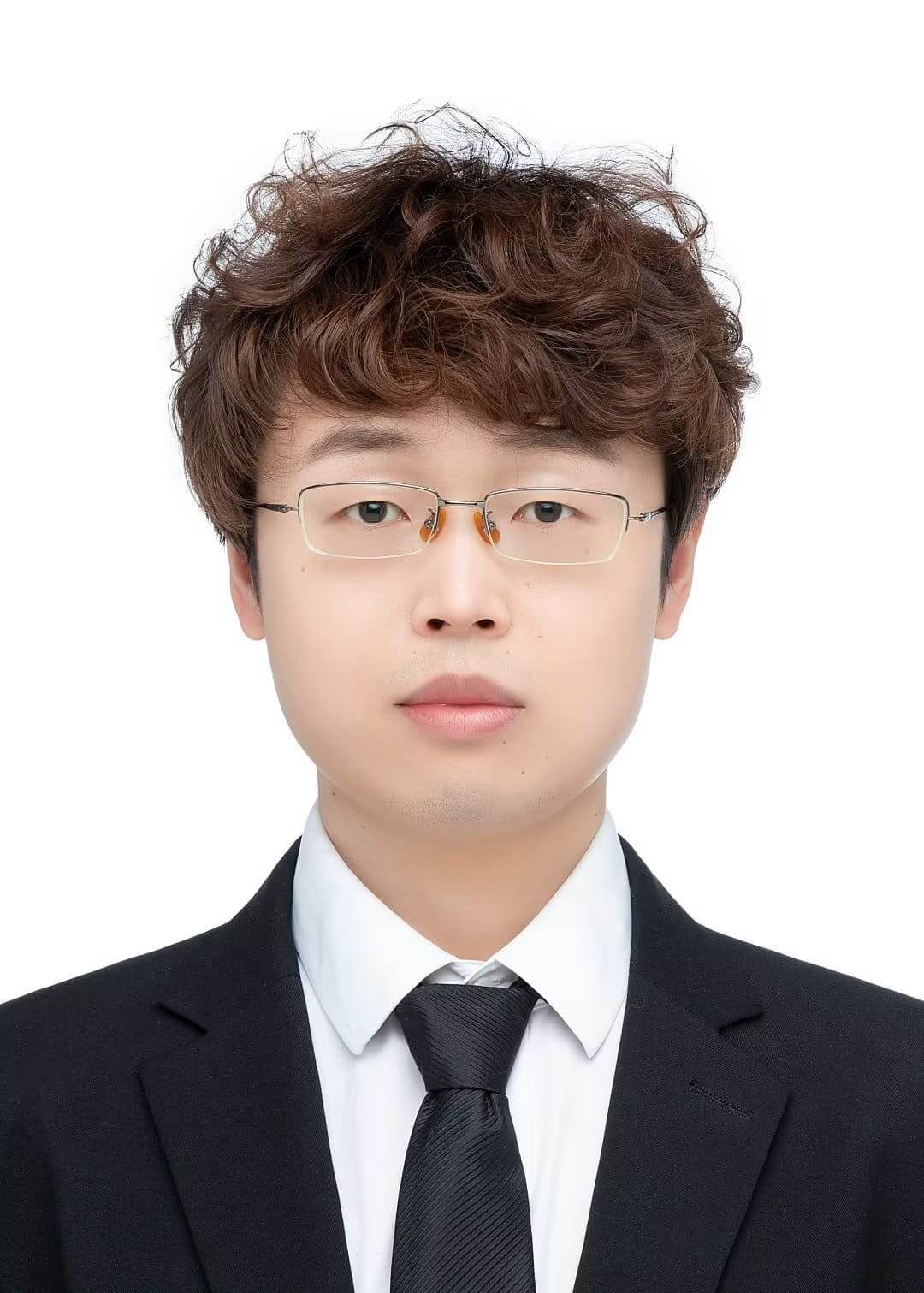}}]{Jing Li}
received the Ph.D. degree in computer science and technology at the College of Intelligence and Computing, Tianjin University, Tianjin, China, in June 2023. He is currently a Lecturer with the School of Computer Science and Engineering, Tianjin University of Technology, Tianjin, China. His research interests lie in the realm of machine learning and computer vision, with a distinct emphasis on developing models and algorithms tailored for real-world scenarios where the abilities to generalize and remain robust are of paramount importance. These scenarios include domain adaptation, open-set recognition, and more.
\end{IEEEbiography}

\vspace{-8 mm}
\begin{IEEEbiography}[{\includegraphics[width=1.0in]{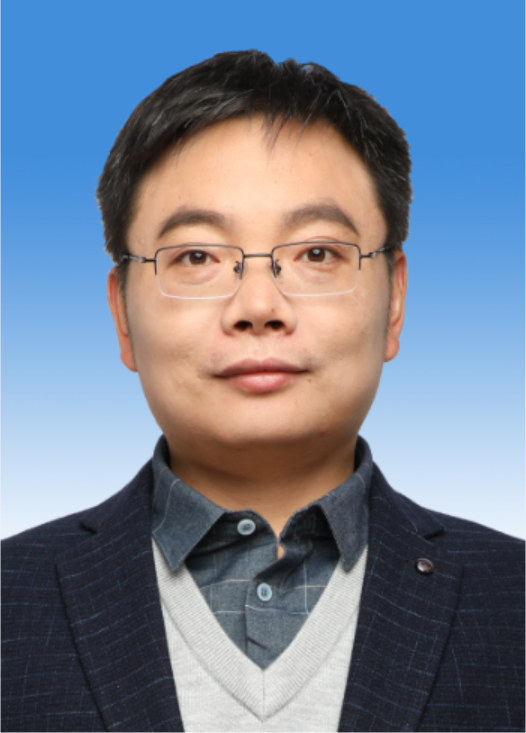}}]{Wanli Xue}
received the Ph.D. degree in computer science and technology at the College of Intelligence and Computing, Tianjin University, Tianjin, China, in 2019. He is currently a Professor with the School of Computer Science and Engineering, Tianjin University of Technology, Tianjin, China. 
His research interests mainly lie in computer vision, including visual tracking and sign language recognition. As the first author or corresponding author, he has published more than 30 papers in prestigious journals and conferences.
\end{IEEEbiography}

\vspace{-10 mm}
\begin{IEEEbiography}[{\includegraphics[width=1.0in]{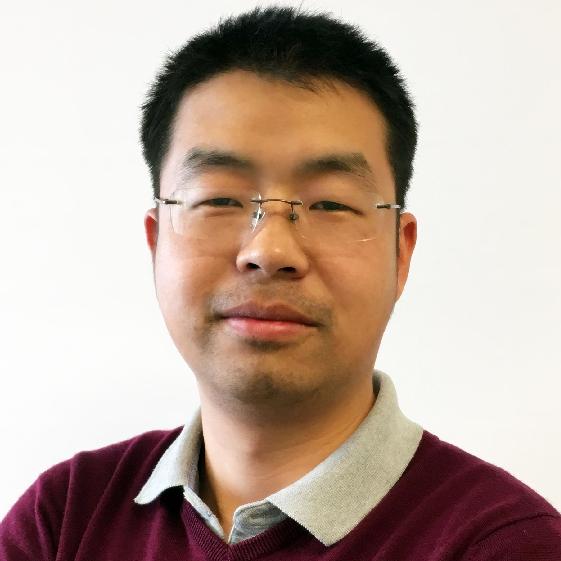}}]{Xu Cheng} (Senior Member, IEEE) received the Ph.D. degree in engineering from the Department of Ocean Operations and Civil Engineering, Intelligent Systems Laboratory, Norwegian University of Science and Technology, Ålesund, Norway, in June 2020.  
He is currently a Professor with Tianjin University of Technology, Tianjin, China. 
He has published more than 60 articles as first and co-author in his research interests, including data analysis and artificial intelligence in maritime operations.
\end{IEEEbiography}

\vspace{-8 mm}
\begin{IEEEbiography}[{\includegraphics[width=1.0in]{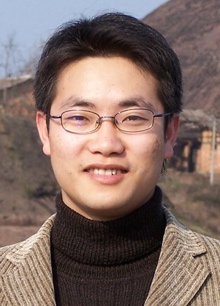}}]{Jianhua Zhang} (Senior Member, IEEE) received the Ph.D. degree from the University of Hamburg, Hamburg, Germany, in 2012. 
He is currently a Professor with the School of Computer Science and Engineering, Tianjin University of Technology, Tianjin, China. His research interests include SLAM, 3D vision, reinforcement learning, and machine vision.
\end{IEEEbiography}

\vspace{-8 mm}
\begin{IEEEbiography}[{\includegraphics[width=1.0in]{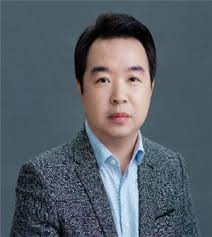}}]{Qinghua Hu}(Senior Member, IEEE) received the B.S., M.S., and Ph.D. degrees from the Harbin Institute of Technology, Harbin, China, in 1999, 2002, and 2008, respectively. He is currently a Peiyang Chair Professor, the Dean of the School of Artificial Intelligence, and the Vice Dean of the Graduate School, Tianjin University, Tianjin, China. 
He has been researching uncertainty modeling in big data, machine learning with multi-modality data, and intelligent systems.
\end{IEEEbiography}

\vspace{-8 mm}
\begin{IEEEbiography}[{\includegraphics[width=1.0in]{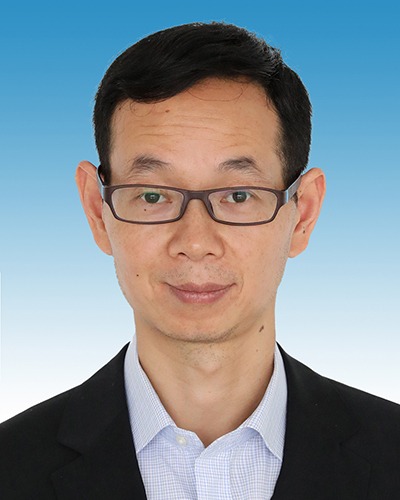}}]{Shengyong Chen} (Senior Member, IEEE) received the Ph.D. degree from City University of Hong Kong in 2003. He worked as a guest researcher at University of Hamburg, Germany, where he received a fellowship from the Alexander von Humboldt Foundation in 2006. He is currently a Professor and the President of Tianjin University of Technology. He is an IET Fellow and an IEEE senior member. His research interests include machine vision and robotics. He has published over 400 scientific papers, including 100 in IEEE Transactions and 5 Best Paper Awards from IEEE.  
\end{IEEEbiography}
\end{document}